%% file: main.tex
\ifdefined\pdfcompresslevel
\fi
\ifdefined\pdfobjcompresslevel
\fi
\ifdefined\pdfminorversion
\fi
\PassOptionsToPackage{table}{xcolor}
\documentclass[]{youtu}
\input{cadflow-compat}
\usepackage{amsmath,amssymb,amsfonts}
\usepackage{graphicx}
\usepackage{booktabs}
\usepackage{multirow}
\usepackage{tabularx}
\usepackage{float}
\usepackage{tikz}
\usepackage{adjustbox}
\usepackage{placeins}
\usepackage{needspace}
\definecolor{ChartBlue}{HTML}{80C4E4}
\definecolor{ChartLine}{HTML}{4E93BD}
\definecolor{ChartAblated}{HTML}{C56A43}
\definecolor{ChartSand}{HTML}{F2CA7D}
\definecolor{ChartTeal}{HTML}{16877D}
\definecolor{ChartInk}{HTML}{222222}
\definecolor{ChartMuted}{HTML}{666666}
\definecolor{ChartGrid}{HTML}{E8E8E8}
\definecolor{ChartLink}{HTML}{BBC7D5}
\definecolor{TableAccent}{HTML}{F3F4F6}

\definecolor{TableDrop}{HTML}{C0392B}
\usepackage{algorithm}
\usepackage{algorithmic}
\usepackage{xcolor}
\usepackage{subcaption}

\usepackage{hyperref}
\usepackage{url}

\newcommand{\ZZ}{\mathbb{Z}}
\newsavebox{\compacttablebox}

\makeatletter
\newcommand{\appendixcontentsline}[3]{\begingroup
  \ifnum#1=3\relax
    \def\appendixnumberwidth{3.5em}\else
    \def\appendixnumberwidth{2.8em}\fi
  \ifnum#1=1\relax
    \bfseries
    \@dottedtocline{#1}{#2}{\appendixnumberwidth}{\hyperref[#3]{\numberline{\ref*{#3}}\nameref*{#3}}}{\hyperref[#3]{\pageref*{#3}}}\else
    \@dottedtocline{#1}{#2}{\appendixnumberwidth}{\hyperref[#3]{\textcolor{ChartInk}{\numberline{\ref*{#3}}\nameref*{#3}}}}{\hyperref[#3]{\pageref*{#3}}}\fi
  \endgroup
}
\makeatother

\title{Can One Hear Geometry?}

\subtitle{Gauge-Invariant Spectral Descriptors for Text-to-B-rep Retrieval}

\author{Hongzhu~Yi$^{1,*}$, Hao~Liu$^{1,*}$, Tao~Yu$^{2}$, Tiankun~Yang$^{1}$, Yujia~Yang$^{1}$,\\
  Zhenghao~Zhang$^{1}$, Tianyu~Zong$^{1}$, Zhenyu~Guan$^{1}$, Bingkang~Shi$^{3}$,\\
  Haopeng~Jin$^{4}$, Yuanxiang~Wang$^{1}$, Zhongtian~Luo$^{4}$,  Kun Tang$^{1}$, Tong Li$^{5}$, Jungang~Xu$^{1,\dagger}$}
\affiliation{$^{1}$University of Chinese Academy of Sciences\\
  $^{2}$Institute of Automation, Chinese Academy of Sciences\\
  $^{3}$Institute of Information Engineering, Chinese Academy of Sciences\\
  $^{4}$Tencent\\
  $^{5}$Washington University, Saint Louis}
\renewcommand{\contributionlist}{\begin{center}\footnotesize
    $^{*}$Equal contribution.\qquad $^{\dagger}$Corresponding author.
  \end{center}}
\hypersetup{pdfauthor={Hongzhu Yi, Hao Liu, Tao Yu, Tiankun Yang, Yujia Yang, Zhenghao Zhang, Tianyu Zong, Zhenyu Guan, Bingkang Shi, Haopeng Jin, Yuanxiang Wang, Zhongtian Luo, Kun Tang, Tong Li, Jungang Xu}}
\hypersetup{pdftitle={Can One Hear Geometry? Gauge-Invariant Spectral Descriptors for Text-to-B-rep Retrieval},pdfsubject={Text-to-B-rep Retrieval},pdfkeywords={text-to-CAD retrieval, B-rep, spectral geometry, geometric invariance}}
\runningtitle{Text-to-B-rep Retrieval}
\reportid{Text-to-B-rep Retrieval}
\reportversion{1.0}
\reportstatus{Technical Report}
\date{October 2026}
\metadata[Resources]{\href{https://github.com/hao2530993905-eng/Text-to-B-rep-Retrive}{Code (GitHub)}\quad\href{https://huggingface.co/hongzhuyi/qwen3-embedding-4b-cad-retrieval-lora}{Model (Hugging Face)}}

\reportabstract{\input{section/abstract}}

\input{report-layout}
\begin{document}
\maketitle
\clearpage
\section*{Contents}
\begingroup
\setcounter{tocdepth}{2}
\setlength{\parskip}{0pt}
\small
\setlength{\baselineskip}{11pt}
\tableofcontents
\endgroup
\clearpage

\begin{figure}[htbp]
\centering
\includegraphics[width=\linewidth]{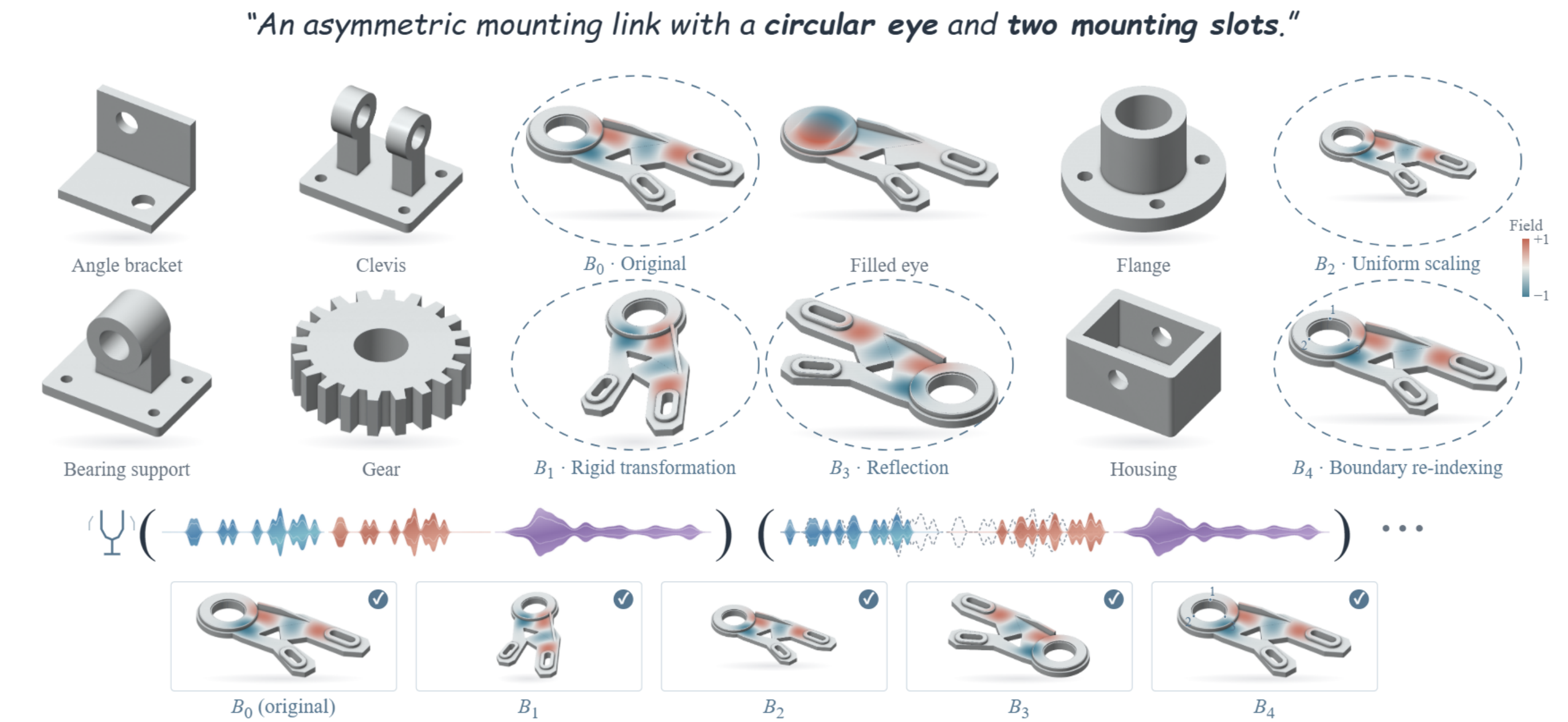}
\caption{
\textbf{Different B-reps, the same geometric identity.}
A single geometric shape can appear in multiple B-rep representations,
yet its underlying geometric identity remains unchanged.
Our gauge-invariant spectral descriptors capture this shared identity
across variations in pose, scale, reflection, and boundary encoding,
enabling geometrically equivalent models to be retrieved together
from a single natural-language query.
Meanwhile, genuinely different shapes, such as the filled-eye variant,
remain distinguishable.
The spectral fields and waveforms visually illustrate the shared
geometric signatures.
}
\label{fig:teaser}
\end{figure}

\input{section/intro}
\input{section/related-work}
\input{section/method}

\input{section/experiments}
\FloatBarrier
\input{section/conclusion}
\FloatBarrier

\phantomsection
\addcontentsline{toc}{section}{References}
\begingroup
\small
\setlength{\bibsep}{3pt}
\bibliography{references}
\endgroup

\clearpage

\appendix
\addtocontents{toc}{\protect\setcounter{tocdepth}{1}}
\section*{Appendix}
\phantomsection
\addcontentsline{toc}{section}{Appendix}

\input{section/appendix}

\end{document}

%% file: cadflow-compat.tex
\usepackage{mathpazo}
\usepackage[numbers,sort&compress]{natbib}
\usepackage{enumitem}
\makeatletter
\newcommand{\subtitle}[1]{\g@addto@macro\titlelist{\begin{center}{\large #1}\end{center}}}
\makeatother
\newcommand{\runningtitle}[1]{\fancyhead[R]{\tencentsanswseven{#1}}}
\newcommand{\reportid}[1]{\metadata[Report]{#1}}
\newcommand{\reportversion}[1]{\metadata[Version]{#1}}
\newcommand{\reportstatus}[1]{\papertype{#1}}
\newcommand{\reportabstract}[1]{\abstract{#1}}

%% file: section/abstract.tex
Text-to-3D solid retrieval faces a fundamental representation gap: while natural language describes intrinsic shapes, Boundary Representation (B-rep) entangles precise geometry with arbitrary variations in pose, scale, and boundary encoding conventions.
Existing methods either discretize B-reps at the cost of analytic precision, or learn features from topological graphs without explicit invariance guarantees.
We propose \textbf{SpectraBRep} (\textit{Spectral Geometry for Text-to-B-rep Retrieval}), a framework based on geometric invariance and spectral descriptors that mathematically eliminates representation ambiguities prior to cross-modal alignment.
Formalizing translation, rotation, reflection, uniform scaling, and contour re-indexing via group actions, we exploit the planar-profile-and-extrusion structure of engineering solids to construct a unified spectral descriptor across cross-sectional regions, contour boundaries, and extrusion depths.
On public benchmarks, our method achieves state-of-the-art recall performance across all evaluation metrics.

%% file: report-layout.tex
\renewcommand{\mymaketitle}{\setlength{\parindent}{0pt}\setlength{\parskip}{0.15cm}\begingroup
    \setlength{\parskip}{0pt}\begin{tcolorbox}[
      colback=youtuLightBlue,
      left=0.5cm,right=0.5cm,top=0.5cm,bottom=0.5cm,
      arc=10pt,enhanced,frame hidden,before skip=0pt,
      grow to left by=1.5pt,grow to right by=1.5pt
    ]
      {\setstretch{1.618}{\color{youtuBlue}\titlelist}\par}\vskip 0.15cm
      \authorlist\par
      \affiliationlist\par
      \contributionlist\par
      \vskip 0.25cm
      \ifdefempty{\metadatalist}{}{\metadatalist\par}\end{tcolorbox}
  \endgroup
  \section*{Abstract}
  \abstractlist\par
  \FloatBarrier
}
\hypersetup{linktoc=all}

\makeatletter
\renewcommand{\tableofcontents}{\@starttoc{toc}}
\makeatother

%% file: section/intro.tex
\FloatBarrier
\section{Introduction}
\label{sec:introduction}

Text-to-3D retrieval aims to identify three-dimensional objects whose semantics and geometric structures match a natural-language description. As collections of 3D design assets continue to grow, this task is gradually extending from generic object retrieval to engineering model retrieval~\citep{seff2020sketchgraphs,chang2015shapenet,koch2019abc,willis2021fusion}. In engineering design, Boundary Representation (B-rep) is the de facto standard for solid geometry. B-rep describes object boundaries through parametric curves, parametric surfaces, and their topological relations, preserving analytic geometric information that is difficult to represent explicitly with discrete representations such as point clouds, voxels, and meshes~\citep{requicha1980solids,jayaraman2021uvnet,lambourne2021brepnet}. Text-based retrieval of engineering models therefore requires establishing a correspondence between natural language and such precise geometric representations.

\begin{figure}[t]
    \centering
    \includegraphics[width=\columnwidth]{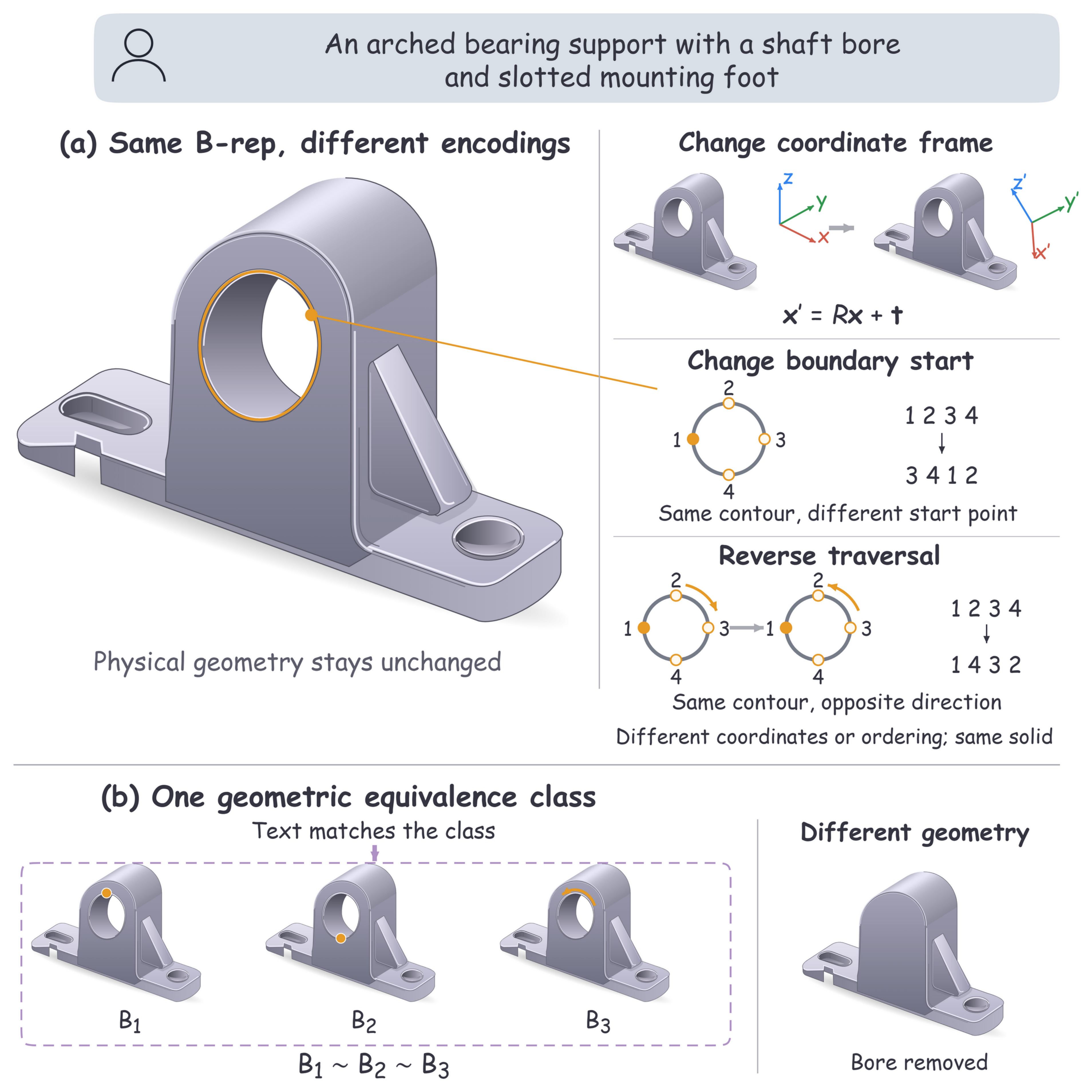}
    \caption{\textbf{Geometry is shared across B-rep encodings.} (a) Changing the coordinate frame, contour starting point, or traversal direction changes the numerical encoding while preserving the solid. (b) The equivalent instances $B_1\sim B_2\sim B_3$ describe the same solid despite different encodings. Removing the bore changes the geometry and no longer matches the description.}
    \label{fig:intro_equivalence}
\end{figure}
This retrieval task is fundamentally challenged by an \textit{asymmetry of abstraction} between natural language and B-rep data. Natural language is inherently invariant to representation artifacts: a prompt like ``a hexagonal nut with a central hole'' describes an intrinsic shape, completely agnostic to its spatial pose, scale, or how its boundary curves are parameterized. In contrast, B-rep entangles intrinsic geometry with extrinsic representation choices---the same geometric entity can yield drastically different numerical descriptions due to variations in coordinate frames, surface parameterizations, and contour starting points (Fig.~\ref{fig:intro_equivalence}a). Existing methods typically bypass this by converting B-reps into discrete point clouds~\citep{qi2017pointnet,qi2017pointnetpp,wang2019dgcnn} or meshes~\citep{hanocka2019meshcnn,wu2015shapenets}, which sacrifices analytic precision and introduces sampling sensitivity. Alternatively, graph-based approaches~\citep{jayaraman2021uvnet,lambourne2021brepnet} operate directly on B-rep topology, yet they conflate the underlying geometry with its parametric encoding. Consequently, these networks must expend significant capacity learning invariance to nuisance transformations via data augmentation, rather than focusing on shape discrimination. Therefore, a central challenge lies in extracting features from B-reps and decoupling intrinsic geometry from arbitrary boundary encoding conventions prior to cross-modal alignment.
To bridge this gap, we propose \textbf{SpectraBRep} (\textit{Spectral Geometry for Text-to-B-rep Retrieval}), a framework that formulates B-rep retrieval on \textit{geometric equivalence classes} (Fig.~\ref{fig:intro_equivalence}b). By formalizing shape-preserving transformations (translation, rotation, reflection, uniform scaling, and contour re-indexing) as group actions, we can quotient out representation ambiguities and construct descriptors that are invariant by design. We realize this mathematical quotient through spectral geometry~\citep{reuter2006laplace,sun2009concise,aubry2011wave}, leveraging the fact that the spectrum of differential operators is intrinsically isometric-invariant. Crucially, we observe a structural isomorphism between extruded B-rep geometry and spectral theory: the prevalent ``planar-profile-and-extrusion'' structure in engineering solids naturally corresponds to the \textit{separation of variables} in the 3D Laplacian. This allows us to analytically synthesize 3D solid descriptors by combining the 2D Dirichlet spectrum (capturing global regional shape), the Fourier magnitudes of boundary turning angles (capturing local perimeter transitions), and the 1D extrusion depth. Without resorting to 3D meshing or data-driven invariance learning, our spectral descriptor compactly unifies regional, boundary, and extrusion structures using spectral invariance principles.
\mbox{SpectraBRep} uses the invariant descriptor as the geometric input to a dual encoder trained on paired text and B-rep models. A residual MLP embeds the descriptor, while Qwen3-Embedding-0.6B encodes the description and is adapted with LoRA. A symmetric contrastive objective aligns each description with its paired B-rep model. We evaluate this compact text encoder on Text2CAD~\cite{khan2024text2cad} and MMCAD-B, separating the construction of invariant geometry features from learned cross-modal matching.

\subsection{Contributions}
\begin{itemize}
    \item \textbf{A Quotient-Space Perspective on B-rep Representation.} We identify that a core bottleneck in Text-to-B-rep retrieval is the entanglement of intrinsic geometry with extrinsic boundary encoding conventions. By formulating geometric invariance via group actions and equivalence classes, we shift the paradigm from data-driven invariance learning to representations constructed from spectral invariants.
    \item \textbf{Analytic Spectral Descriptors for B-rep Solids.} Exploiting the parallel between extruded solid geometry and separation of variables, we construct a unified spectral descriptor combining Dirichlet eigenvalues and boundary Fourier magnitudes. It captures regional, boundary, and extrusion geometry directly from B-reps without requiring volumetric meshing.
\end{itemize}

%% file: section/related-work.tex
\FloatBarrier
\section{Related work}
\label{sec:related_work}

\paragraph{Text-to-3D retrieval.}
Cross-modal retrieval between text and 3D shapes has been explored through point clouds~\citep{xue2023ulip,zhou2024uni3d,chen2023text2shape}, meshes, and rendered images~\citep{su2015multiview,zhang2022pointclip}.
ShapeGlot~\citep{achlioptas2019shapeglot} learns joint embeddings between language and 3D shapes via contrastive objectives.
Text2Shape~\citep{chen2023text2shape} aligns text with colored point clouds.
More recent methods leverage pretrained vision-language models~\citep{zhu2023pointclipv2,radford2021learning,zhang2022pointclip,xue2024ulip2,zhou2024uni3d}: ULIP~\citep{xue2023ulip} and OpenShape~\citep{liu2024openshape} align point-cloud encoders with CLIP embeddings.
However, these methods target mesh or point-cloud representations and do not address parametric CAD models with their construction-sequence structure.

\paragraph{CAD representation learning.}
DeepCAD~\citep{wu2021deepcad} pioneered autoregressive generation of CAD construction sequences.
SkexGen~\citep{xu2022skexgen} disentangles sketch and extrusion generation.
CAD-SIGNet~\citep{khan2024cadsignet} learns signatures over construction graphs.
Text2CAD~\citep{khan2024text2cad} provides multi-level text descriptions for DeepCAD and generates parametric CAD construction sequences from text.
CADIR~\citep{liu2026cadir} introduces an executable CAD intermediate representation and supports text- and image-based retrieval of construction graphs.

Published results for BRepCLIP~\citep{usama2026brepclip}, Duoduo CLIP~\citep{lee2025duoduoclip}, CLIP-GS~\citep{jiao2025clipgs}, CADGCL~\citep{qin2025cadgcl}, and BRT~\citep{zou2025brt} are summarized in Appendix~\ref{app:published_baselines}.

\paragraph{Broader representation learning and multimodal systems.}
Language-grounded retrieval connects to a broader literature on representation objectives, multimodal understanding, and model-based interaction.
Norm-aware contrastive learning in TNCSE, JTCSE, TNCME, and L2Dir studies how embedding magnitudes complement directional alignment in sentence and multimodal representations~\citep{zong2025tncse,zong2025jtcse,zong2026tncme,zong2026l2dir}.
Related representation objectives include dynamic graph reconstruction for incomplete multi-view clustering~\citep{zhang2025dgimvcm} and spectral distribution matching for dataset distillation~\citep{wu2026spectraldistillation}, while LCGC and ABM address class imbalance through prediction debiasing and feature-aware calibration~\citep{xing2025lcgc,yi2025abm}.
For visual sequences, HIMMEL separates semantic and motion encoding~\citep{jin2026himmel}, Delta-JEPA learns action-sensitive latent dynamics~\citep{zhang2026deltajepa}, and OPIS evaluates the preservation of object identity and structure in video world models~\citep{wang2026opis}.
Complementary model ensembles support video question answering~\citep{xie2025foureyes,xie2025teamofone}, and emotion recognition has been studied through human-preference-aligned experts, zero-shot ensembles, and reliable reasoning priors~\citep{xie2025more,xie2025zeroes,wang2025reliablepriors}.
Instruction-guided image editing raises related questions of semantic alignment and content preservation, studied through residual line search~\citep{yi2026manifold} and evaluated by Omni IIE Bench and VDE Bench~\citep{yang2026omniiie,yi2026vde}.
Open-domain retrieval extends language grounding to shot search, memory-based video localization, and audio-driven cross-modal evidence gathering~\citep{yu2026shotfinder,yu2026rvms,yu2026omnideepsearch}.
Agent systems further connect external information to task execution through browser interaction~\citep{yu2025browseragent}, structured academic presentation generation~\citep{yu2026paperx}, and scientific workflows surveyed in the autonomous-research literature~\citep{wang2026hitchhiker}.
Their execution is studied through cost-aware model routing~\citep{yu2026scoperouter}, role-specialized collaboration~\citep{yu2026entcollabbench}, and skill learning under noisy experience~\citep{yu2026rsebench}.
Experience-replay and partial-reasoning formulations study efficient reasoning optimization~\citep{yi2026poer,yi2026rpo}, while MR-ALIGN targets factuality through reasoning-process alignment~\citep{wang2026mralign}.
Complementary evaluations examine search-grounded video misinformation detection~\citep{yu2026evidbench} and social biases in language-model-based game characters~\citep{shi2026fairgamer}.
Within this broader context, our focus is the geometric identity preserved across equivalent B-rep encodings and its alignment with language for CAD retrieval.

%% file: section/method.tex
\FloatBarrier
\section{Preliminaries}
\label{sec:method}

\begin{figure}[!htbp]
\centering
\includegraphics[width=\linewidth]{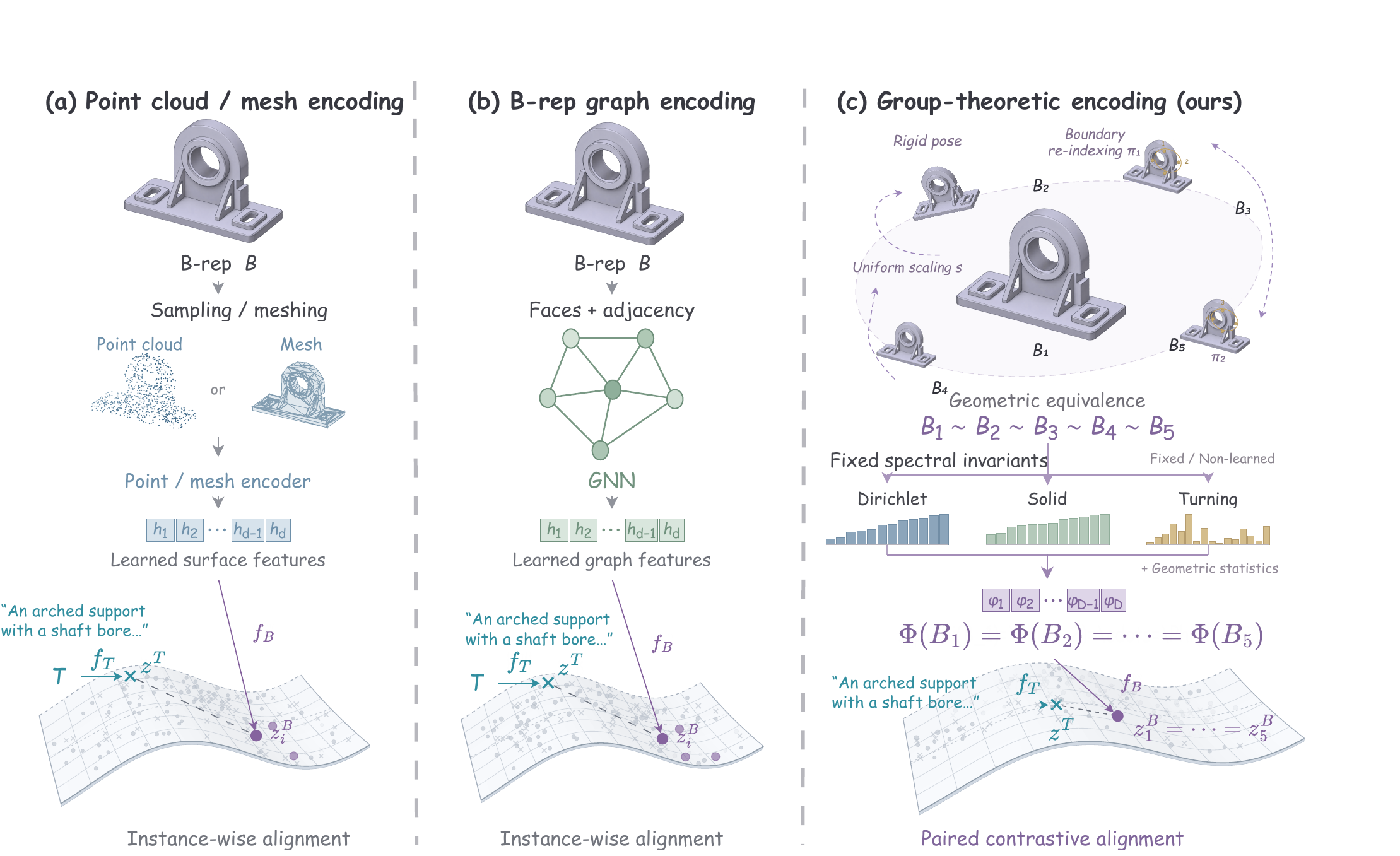}
\caption{\textbf{Comparison of text-to-B-rep retrieval pipelines.} (a) Point-cloud and mesh pipelines sample or tessellate the B-rep, learn surface features, and use instance-wise text--geometry alignment. (b) Graph pipelines encode faces and their adjacency with a GNN, again aligning individual instances. (c) Our group-theoretic pipeline first maps rigidly transformed, uniformly scaled, reflected, or boundary-reindexed B-reps to fixed, non-learned spectral invariants and geometric statistics. A trainable geometry encoder then embeds the shared descriptor, which is aligned with the paired text through a symmetric contrastive objective.}
\label{fig:method_overview}
\end{figure}

\subsection{Problem formulation}
\label{sec:setup}
Given a text query $T_i$ and a construction sequence $X_i$, the B-rep geometry is reconstructed via:
\begin{equation}
\mathcal B_i = \mathcal R(X_i),
\label{eq:reconstruction}
\end{equation}

where $\mathcal{R}$ is the reconstruction mapping~\citep{wu2021deepcad,willis2021fusion}. Due to the asymmetry between abstract natural language and raw B-rep pose/parameter noise, direct alignment is inefficient. In contrast to the instance-wise point/mesh and graph pipelines in Figure~\ref{fig:method_overview}(a,b), our pipeline applies a deterministic quotient map before the trainable geometry encoder and paired text--B-rep alignment (Figure~\ref{fig:method_overview}c):
\begin{equation}
\Phi: \mathcal B_i \mapsto \boldsymbol d_i,
\label{eq:quotient_map}
\end{equation}

which collapses equivalent geometries into an invariant descriptor $\boldsymbol d_i$, enabling encoders to focus purely on semantic-shape alignment.

\subsection{Structural priors and equivalence relations}
\label{sec:brep_reconstruction}

\paragraph{Equivalence transformations factored out by $\Phi$.}
To ensure invariance, $\Phi$ must factor out extrinsic similarity transformations and intrinsic recording conventions (contour starts, traversal directions, and unit indices)~\citep{cohen2016equivariant}. Denoting this equivalence as $\mathcal B'\sim\mathcal B$, we use the following relation as an ideal invariance criterion for descriptor design:
\begin{equation}
\mathcal B'\sim\mathcal B
\quad\Longrightarrow\quad
\Phi(\mathcal B')=\Phi(\mathcal B),
\label{eq:invariant_requirement}
\end{equation}

while remaining sensitive to substantive geometric changes (Appendices~\ref{app:geometry} and~\ref{app:gauge}).

\begin{figure}[!htbp]
\centering
\includegraphics[width=\columnwidth]{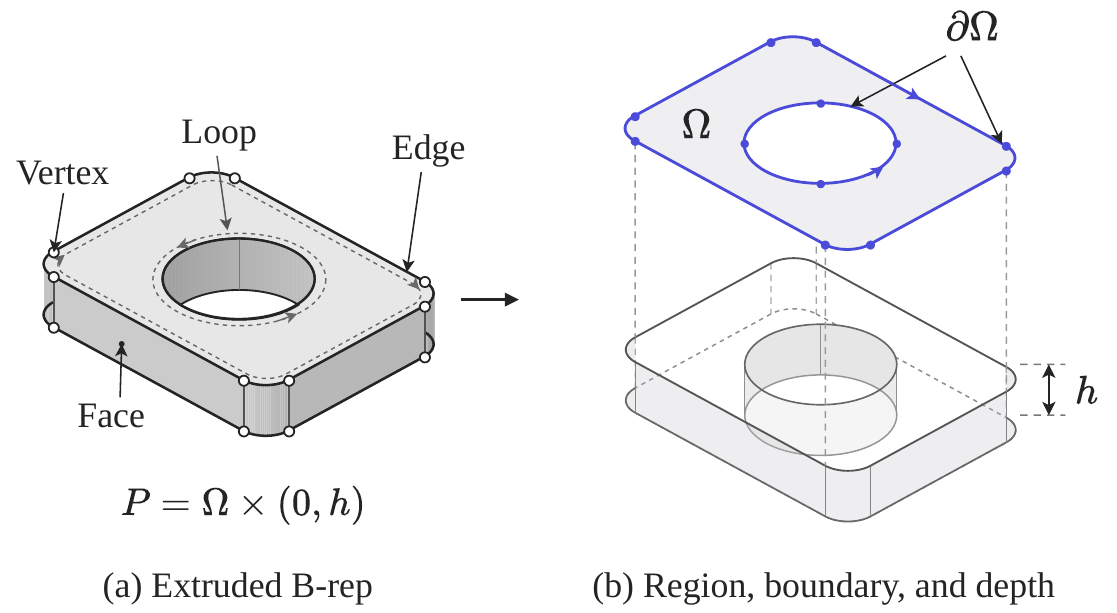}
\caption{\textbf{Product decomposition of a straight-extrusion unit.} (a) A rounded B-rep solid with a through-hole, showing faces, edges, vertices, and oriented boundary loops (dashed arrows). (b) Its planar region $\Omega$, boundary $\partial\Omega$, and extrusion depth $h$; $\partial\Omega$ sweeps along thickness $h$ to form the lateral and inner-hole walls.}
\label{fig:brep_geometry}
\end{figure}

\paragraph{Product structure of extruded B-reps.}
Exploiting the sketch-and-extrusion paradigm, engineering B-reps naturally admit a Cartesian product decomposition for each straight-extrusion unit:
\begin{equation}
P=\Omega\times(0,h),
\label{eq:extruded_solid}
\end{equation}

where $\Omega \subset \mathbb{R}^2$ is the cross-section and $h>0$ is the depth (Figure~\ref{fig:brep_geometry}). This 2D-1D product structure decomposes the 3D invariance problem into tractable regional and boundary subproblems (Appendix~\ref{app:geometry}).

As illustrated in the Figure ~\ref{fig:brep_geometry}, a B-rep solid can be decomposed into the product of these three features, while any complex B-rep formed by the combination of multiple B-rep units can in turn be expressed as the sum of these simple B-rep features. Interestingly, since both products and sums are permutation-invariant operations, we can leverage this property to construct order-invariant features. Our method is presented as follows.

\section{SpectraBRep}

\subsection{Constructing orthogonal spectral invariants}
\label{sec:descriptor}                           
Under the decomposition $P=\Omega\times(0,h)$ in Eq.~\eqref{eq:extruded_solid}, the geometry of an extrusion unit is determined by its planar region $\Omega$, extrusion depth $h$, and closed boundary $\partial\Omega$. \mbox{SpectraBRep} constructs spectral features from these three components in turn. We first extract coordinate-independent cross-sectional features from $\Omega$, then incorporate the 3D thickness ratio using $h$, next capture local corner geometry while eliminating contour-indexing variations on $\partial\Omega$, and finally aggregate unit features into the model-level descriptor $\Phi(\mathcal B_i)$.

\begin{figure}[t]
\centering
\includegraphics[width=\textwidth]{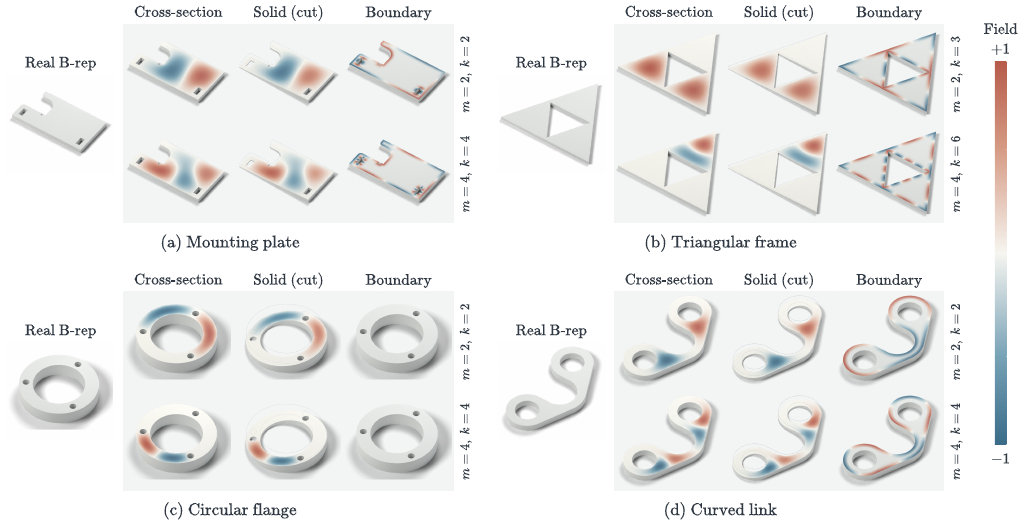}
\caption{\textbf{Spectral components on real B-reps.} Four MMCAD-B parts are shown with regional Dirichlet modes on their top faces, solid modes $u_m(x,y)\sin(\pi z/h)$ on oblique interior cuts, and turning harmonics along their boundaries. The two rows use $m=2,4$ and $k=2,4$; the triangular frame uses $k=3,6$. Colors indicate signed amplitudes normalized per field or boundary loop; zero harmonics remain neutral. Boundary panels include outer and hole contours, while the descriptor retains outer-contour Fourier magnitudes.}
\label{fig:spectral_teaser}
\end{figure}

Figure~\ref{fig:spectral_teaser} visualizes these three components on four real B-reps, showing regional oscillations, depth-dependent solid modes, and contour-localized turning harmonics.

\paragraph{Regional Dirichlet spectrum.}
\label{sec:dirichlet}

We first consider the planar cross section $\Omega$ (including internal holes) and extract shape features that do not change under 2D coordinate translation or rotation. The eigenvalues of the Laplace operator are invariant under isometries, meaning that the resonant frequencies of a domain depend only on its intrinsic geometry rather than its placement in the plane~\citep{kac1966can,peinecke2007fingerprints}. Accordingly, we solve the Dirichlet Laplacian eigenvalue problem on $\Omega$:
\begin{equation}
\begin{gathered}
-\Delta u_k=\mu_k u_k\quad\text{in }\Omega,\qquad u_k=0\quad\text{on }\partial\Omega,\\
0<\mu_1\leq\mu_2\leq\cdots.
\end{gathered}
\label{eq:dirichlet}
\end{equation}

The eigenvalue sequence $\{\mu_k\}$ is invariant to planar translation, rotation, and reflection~\citep{reuter2006laplace}. Since scaling the domain uniformly by a factor $s$ scales $\mu_k$ by $s^{-2}$, multiplying $\mu_k$ by the cross-sectional area $|\Omega|$ removes the scale factor and gives a dimensionless 2D regional spectrum:
\begin{equation}
    \widehat{\mu}_k = \mu_k |\Omega|.
    \label{eq:normalized_planar_spectrum}
\end{equation}

\paragraph{Extrusion spectrum.}
\label{sec:extrusion_spectrum}

The 2D regional spectrum $\widehat{\mu}_k$ describes only the base cross section $\Omega$ and cannot distinguish a thin washer from a long cylinder with the same base. Solving for 3D eigenvalues directly on $P$ via tetrahedral meshing would introduce high computational cost and volumetric discretization error. Because $P=\Omega\times(0,h)$ in Eq.~\eqref{eq:extruded_solid} is a Cartesian product, the 3D Dirichlet Laplacian admits exact \textit{separation of variables} ($\Delta_P = \Delta_\Omega + \partial_{zz}$). Thus, without 3D meshing, we can combine the 2D eigenvalues $\mu_m$ with the depth $h$ algebraically to obtain the exact 3D eigenvalues of the extrusion unit:
\begin{equation}
    \lambda_{m,n}
    = \mu_m + \left(\frac{n\pi}{h}\right)^2,
    \qquad m,n \geq 1.
    \label{eq:extrusion_spectrum}
\end{equation}

Normalizing by the 3D volume $|P|=|\Omega|h$ removes absolute scale:
\begin{equation}
    \widehat{\lambda}_{m,n}
    = \lambda_{m,n}|P|^{2/3},
    \label{eq:normalized_extrusion}
\end{equation}

so that the normalized extrusion spectrum $\widehat{\lambda}_{m,n}$ captures the 3D aspect ratio between cross-sectional size and extrusion depth (relative depth $h/\sqrt{|\Omega|}$; see Appendix~\ref{app:spectral_invariance}).

\paragraph{Boundary turning Fourier spectrum.}
\label{sec:turning}

To capture local high-frequency geometry and eliminate contour-recording variations, we compute the signed turning-angle sequence along perimeter $L$~\citep{arkin1991efficiently,zahn1972fourier,kuhl1982elliptic,persoon1977fourier} and apply a 1D discrete Fourier transform:
\begin{equation}
    F_k
    = \sum_{j=1}^{N_{\partial}}
    \Delta\theta_j
    \exp\left(-\frac{2\pi\mathrm{i}k s_j}{L}\right).
    \label{eq:turning_spectrum}
\end{equation}

\begin{figure}[t]
    \centering
    \includegraphics[width=\columnwidth]{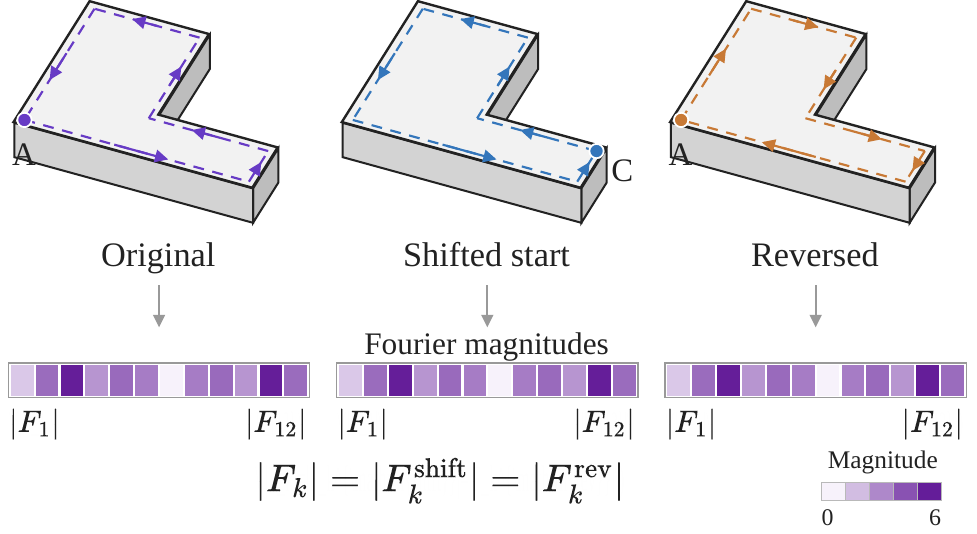}
    \caption{\textbf{Invariance of the boundary spectrum to recording conventions.} Changing the contour starting vertex from $A$ to $C$ or reversing traversal direction only changes phase or conjugates coefficients in the frequency domain; the Fourier magnitudes $|F_k|$ remain identical.}
    \label{fig:boundary_invariance}
\end{figure}

Taking the magnitude $|F_k|$ eliminates start-point, direction, and reflection effects while preserving local corners and arcs (Figure~\ref{fig:boundary_invariance}; Appendix~\ref{app:spectral_invariance}).

\paragraph{Numerical descriptor and assembly.}

We extract extrusion units from the sketch--extrusion records
$X_i$ and their unit indices. Each unit contributes
$K_r$ regional spectral features,
$K_e$ depth-combined spectral features,
$K_b$ outer-boundary Fourier magnitudes, and
$K_a$ geometric and record-level attributes.
The resulting local descriptor
$\mathbf v_u \in \mathbb R^{D_l}$ has dimension
$D_l=K_r+K_e+K_b+K_a$.

We concatenate a global block
$\mathbf g\in\mathbb R^{D_g}$ with $U$ local descriptors:
\[
\boldsymbol d_i=
[\mathbf g;\mathbf v_1;\cdots;\mathbf v_U]
\in\mathbb R^D,
\qquad D=D_g+UD_l.
\]

Units are stably sorted by decreasing representative
outer-loop area; missing slots are zero-filled, and the
global block aggregates all valid units, including those
beyond the retained $U$ slots.
The extrusion block combines the regional spectrum with relative
extrusion depth according to Eq.~\eqref{eq:normalized_extrusion}.
Appendix~\ref{app:descriptor_layout} specifies the
descriptor dimensions, numerical transforms, tie rules,
and standardization procedures.

\subsection{Cross-modal contrastive alignment}
\label{sec:training}

Given the invariant descriptor $\boldsymbol d_i=\Phi(\mathcal B_i)$, we map geometry and text into a shared $D$-dimensional space using a dual-encoder architecture~\citep{radford2021learning,xue2023ulip,chen2020simclr,wang2020alignment}. Specifically, a residual MLP $f_B$ encodes $\boldsymbol d_i$ into $\boldsymbol z_i^B=\operatorname{Norm}\!\left(f_B(\boldsymbol d_i)\right)$, while a text encoder $f_T$ encodes the text $T_i$ into $\boldsymbol z_i^T=\operatorname{Norm}\!\left(f_T(T_i)\right)$.

Cross-modal similarity is measured by the cosine inner product $S_{ij}=\langle\boldsymbol z_i^T,\boldsymbol z_j^B\rangle$.

We adapt the text encoder with LoRA. For a batch of $N$ paired descriptions and B--rep models, the target of description $T_i$ is its paired model $\mathcal B_i$. With temperature $\tau$, we optimize the symmetric contrastive objective
\begin{equation}
\mathcal L_{\mathrm{con}}
= -\frac{1}{2N}\sum_{i=1}^{N}
\log\frac{e^{2S_{ii}/\tau}}
{\bigl(\sum_j e^{S_{ij}/\tau}\bigr)
 \bigl(\sum_j e^{S_{ji}/\tau}\bigr)}.
\label{eq:contrastive_loss}
\end{equation}

The two sums run over the batch in the text-to-geometry and geometry-to-text directions, respectively. The training objective is $\mathcal L=\mathcal L_{\mathrm{con}}$. At inference, candidate B-rep embeddings are precomputed offline; a query is encoded once and candidates are ranked by cosine similarity.

%% file: section/experiments.tex
\FloatBarrier
\section{Experiments}
\label{sec:experiments}
\suppressfloats[t]
\begin{table}[!htbp]

\centering
\small
\setlength{\tabcolsep}{2.5pt}
\renewcommand{\arraystretch}{1.1}
\begin{adjustbox}{max width=\linewidth}
\begin{tabular}{@{}lc cccc cccc@{}}
\toprule
\multirow{2}{*}{Method} & \multirow{2}{*}{Params} & \multicolumn{4}{c}{\textbf{Text2CAD}} & \multicolumn{4}{c}{\textbf{MMCAD-B}} \\
\cmidrule(lr){3-6}\cmidrule(l){7-10}
& & R@1$\uparrow$ & R@5$\uparrow$ & R@10$\uparrow$ & MedR$\downarrow$ & R@1$\uparrow$ & R@5$\uparrow$ & R@10$\uparrow$ & MedR$\downarrow$ \\
\midrule
CLIP (zero-shot) & 151.3M & \cellcolor{youtuBlue!4.0}0.011$_{\pm 0.002}$ & \cellcolor{youtuBlue!4.0}0.059$_{\pm 0.004}$ & \cellcolor{youtuBlue!4.0}0.091$_{\pm 0.005}$ & \cellcolor{youtuBlue!4.0}188 & \cellcolor{youtuBlue!4.0}0.008$_{\pm 0.001}$ & \cellcolor{youtuBlue!4.0}0.035$_{\pm 0.003}$ & \cellcolor{youtuBlue!4.0}0.072$_{\pm 0.004}$ & \cellcolor{youtuBlue!4.0}215 \\
BLIP-2 (zero-shot) & 3.1B & \cellcolor{youtuBlue!4.1}0.015$_{\pm 0.002}$ & \cellcolor{youtuBlue!4.1}0.065$_{\pm 0.004}$ & \cellcolor{youtuBlue!4.5}0.112$_{\pm 0.006}$ & \cellcolor{youtuBlue!7.9}152 & \cellcolor{youtuBlue!4.2}0.011$_{\pm 0.002}$ & \cellcolor{youtuBlue!4.4}0.048$_{\pm 0.003}$ & \cellcolor{youtuBlue!4.5}0.089$_{\pm 0.005}$ & \cellcolor{youtuBlue!7.7}176 \\
\midrule
TriCoLo & 20.4M & \cellcolor{youtuBlue!8.2}0.143$_{\pm 0.005}$ & \cellcolor{youtuBlue!11.6}0.372$_{\pm 0.008}$ & \cellcolor{youtuBlue!13.4}0.491$_{\pm 0.009}$ & \cellcolor{youtuBlue!22.8}12 & \cellcolor{youtuBlue!9.0}0.098$_{\pm 0.004}$ & \cellcolor{youtuBlue!10.8}0.252$_{\pm 0.006}$ & \cellcolor{youtuBlue!12.3}0.358$_{\pm 0.007}$ & \cellcolor{youtuBlue!22.1}22 \\
RoMa & 12.9M & \cellcolor{youtuBlue!9.4}0.182$_{\pm 0.006}$ & \cellcolor{youtuBlue!13.0}0.428$_{\pm 0.008}$ & \cellcolor{youtuBlue!15.0}0.559$_{\pm 0.008}$ & \cellcolor{youtuBlue!23.1}9 & \cellcolor{youtuBlue!8.6}0.091$_{\pm 0.004}$ & \cellcolor{youtuBlue!10.0}0.224$_{\pm 0.006}$ & \cellcolor{youtuBlue!11.5}0.332$_{\pm 0.008}$ & \cellcolor{youtuBlue!21.7}26 \\
Parts2Words & 16.0M & \cellcolor{youtuBlue!10.6}0.219$_{\pm 0.006}$ & \cellcolor{youtuBlue!14.9}0.507$_{\pm 0.008}$ & \cellcolor{youtuBlue!16.5}0.623$_{\pm 0.007}$ & \cellcolor{youtuBlue!23.6}5 & \cellcolor{youtuBlue!12.3}0.158$_{\pm 0.005}$ & \cellcolor{youtuBlue!14.4}0.364$_{\pm 0.007}$ & \cellcolor{youtuBlue!15.8}0.481$_{\pm 0.008}$ & \cellcolor{youtuBlue!23.1}12 \\
SCA3D & 16.0M & \cellcolor{youtuBlue!12.0}0.263$_{\pm 0.007}$ & \cellcolor{youtuBlue!16.0}0.553$_{\pm 0.008}$ & \cellcolor{youtuBlue!17.6}0.671$_{\pm 0.007}$ & \cellcolor{youtuBlue!23.7}4 & \cellcolor{youtuBlue!10.8}0.132$_{\pm 0.005}$ & \cellcolor{youtuBlue!12.8}0.315$_{\pm 0.007}$ & \cellcolor{youtuBlue!14.4}0.430$_{\pm 0.008}$ & \cellcolor{youtuBlue!22.8}15 \\
BRepCLIP \citep{usama2026brepclip} & 86.0M & \cellcolor{youtuBlue!13.0}0.295$_{\pm 0.007}$ & \cellcolor{youtuBlue!16.7}0.582$_{\pm 0.008}$ & \cellcolor{youtuBlue!18.1}0.694$_{\pm 0.007}$ & \cellcolor{youtuBlue!23.7}4 & \cellcolor{youtuBlue!15.2}0.211$_{\pm 0.006}$ & \cellcolor{youtuBlue!16.6}0.435$_{\pm 0.008}$ & \cellcolor{youtuBlue!18.2}0.562$_{\pm 0.008}$ & \cellcolor{youtuBlue!23.5}7 \\
Duoduo CLIP \citep{lee2025duoduoclip} & 110.0M & \cellcolor{youtuBlue!13.8}0.318$_{\pm 0.007}$ & \cellcolor{youtuBlue!17.0}0.595$_{\pm 0.008}$ & \cellcolor{youtuBlue!18.3}0.703$_{\pm 0.007}$ & \cellcolor{youtuBlue!23.8}3 & \cellcolor{youtuBlue!14.0}0.189$_{\pm 0.006}$ & \cellcolor{youtuBlue!15.6}0.402$_{\pm 0.008}$ & \cellcolor{youtuBlue!17.2}0.528$_{\pm 0.008}$ & \cellcolor{youtuBlue!23.3}9 \\
IMMA & 561.4M & \cellcolor{youtuBlue!14.4}0.338$_{\pm 0.007}$ & \cellcolor{youtuBlue!17.4}0.608$_{\pm 0.008}$ & \cellcolor{youtuBlue!18.5}0.712$_{\pm 0.007}$ & \cellcolor{youtuBlue!23.8}3 & \cellcolor{youtuBlue!14.7}0.202$_{\pm 0.006}$ & \cellcolor{youtuBlue!17.0}0.448$_{\pm 0.008}$ & \cellcolor{youtuBlue!18.7}0.581$_{\pm 0.008}$ & \cellcolor{youtuBlue!23.6}6 \\
CLIP-GS \citep{jiao2025clipgs} & 150.0M & \cellcolor{youtuBlue!15.2}0.362$_{\pm 0.007}$ & \cellcolor{youtuBlue!17.6}0.620$_{\pm 0.007}$ & \cellcolor{youtuBlue!18.7}0.718$_{\pm 0.006}$ & \cellcolor{youtuBlue!23.8}3 & \cellcolor{youtuBlue!17.1}0.245$_{\pm 0.006}$ & \cellcolor{youtuBlue!18.1}0.482$_{\pm 0.008}$ & \cellcolor{youtuBlue!19.0}0.591$_{\pm 0.008}$ & \cellcolor{youtuBlue!23.6}6 \\
CADIR \citep{liu2026cadir} & 4.0B & \cellcolor{youtuBlue!16.1}0.393$_{\pm 0.008}$ & \cellcolor{youtuBlue!18.0}0.634$_{\pm 0.007}$ & \cellcolor{youtuBlue!18.8}0.721$_{\pm 0.006}$ & \cellcolor{youtuBlue!23.9}2 & \cellcolor{youtuBlue!17.4}0.252$_{\pm 0.007}$ & \cellcolor{youtuBlue!19.0}0.511$_{\pm 0.008}$ & \cellcolor{youtuBlue!20.0}0.624$_{\pm 0.007}$ & \cellcolor{youtuBlue!23.7}5 \\
\midrule
Ours (LoRA) & 0.6B$^{\dagger}$ & \cellcolor{youtuBlue!24.0}\textbf{0.640$_{\pm 0.004}$} & \cellcolor{youtuBlue!24.0}\textbf{0.881$_{\pm 0.004}$} & \cellcolor{youtuBlue!24.0}\textbf{0.945$_{\pm 0.005}$} & \cellcolor{youtuBlue!24.0}\textbf{1} & \cellcolor{youtuBlue!24.0}\textbf{0.371$_{\pm 0.008}$} & \cellcolor{youtuBlue!24.0}\textbf{0.669$_{\pm 0.007}$} & \cellcolor{youtuBlue!24.0}\textbf{0.763$_{\pm 0.006}$} & \cellcolor{youtuBlue!24.0}\textbf{2} \\
Ours (Frozen) & 0.6B$^{\dagger}$ & \cellcolor{youtuBlue!14.6}0.344$_{\pm 0.008}$ & \cellcolor{youtuBlue!17.6}0.618$_{\pm 0.005}$ & \cellcolor{youtuBlue!19.0}0.731$_{\pm 0.004}$ & \cellcolor{youtuBlue!23.8}3 & \cellcolor{youtuBlue!20.3}0.304$_{\pm 0.008}$ & \cellcolor{youtuBlue!22.5}0.622$_{\pm 0.007}$ & \cellcolor{youtuBlue!23.1}0.732$_{\pm 0.006}$ & \cellcolor{youtuBlue!23.9}3 \\
\bottomrule
\end{tabular}
\end{adjustbox}
\caption{\textbf{Text-to-CAD retrieval on Text2CAD and MMCAD-B.} Both use 1,000-model pools with Expert and Description queries, respectively. On both datasets, recalls are mean $\pm$ std across ten randomly selected evaluation seeds for candidate-pool sampling. Ours uses Qwen3-Embedding-0.6B; $\dagger$ marks backbone size. Baseline parameter counts are system-level. See Appendix~\ref{app:baseline_settings} for the protocol. Darker blue indicates better performance within each metric.}
\label{tab:main}
\end{table}    

\subsection{Setup}

The main model uses Qwen3-Embedding-0.6B with LoRA. We evaluate text-to-B-rep retrieval framework on Text2CAD and MMCAD-B by ranking CAD models for each natural-language query and report R@1, R@5, R@10, and median rank (MedR) using candidate pools of 1,000 models. Appendix~\ref{app:baseline_settings} describes the baseline implementations, data partitions, input adaptation, training and checkpoint selection, and evaluation and statistical procedures. The optimization settings of our model are provided in Appendix~\ref{app:training_details}.

\subsection{Validation of the complete descriptor}
\label{sec:full_descriptor_stability}
Our method produces a 280-dimensional descriptor; Appendix~\ref{app:descriptor_layout} details its feature composition and layout.
It is designed to eliminate representation
variations that do not alter the underlying geometry while retaining
distinctions across different CAD construction records.
We evaluate these properties by examining descriptor consistency
under geometry-preserving transformations and collisions across
different construction records
(Appendix~\ref{app:full_descriptor_stability}).

Across models passing B-rep consistency checks, global translation, rotation, and reflection leave the descriptors exactly unchanged. Under contour-recording transformations, 99.88\%--100\% of pairs have relative $\ell_2$ errors within 1\%. To evaluate the inputs seen by the geometry encoder, we standardize each descriptor coordinate using its training-set mean and standard deviation, then round it to the nearest integer (Appendix~\ref{app:assembly_conditions}). We call the resulting integer vector a \emph{quantized key}; two keys agree only if all coordinates are identical. Quantized-key agreement ranges from 99.75\% to 100\%. These results indicate that the invariance of individual geometric features is largely preserved in the complete descriptor assembly.

Furthermore, among 487,473 pairs with different construction vectors, exact descriptor equality and quantized-key agreement occur in only 0.0542\% and 0.1061\% of pairs, respectively (Table~\ref{tab:full_descriptor_stability}). Together with the transformation results above, these findings show that the complete descriptor remains stable across the tested equivalent representations while preserving distinctions across different construction records.

We further examine how well the descriptor distinguishes individual models. Among 4,000 randomly sampled test models, it yields 3,315 distinct representations, compared with 3,601 distinct raw construction sequences; the ratio of these counts is approximately 92\%. We also use the target model's own geometric descriptor as the retrieval query, obtaining R@1=0.913, R@5=0.951, and R@10=0.989 (Appendix~\ref{app:validation_details}). This geometry-only test provides the empirical self-retrieval reference used below to interpret text-to-CAD performance.

\subsection{Main retrieval results}
\label{sec:main_results}
All trainable baselines and SpectraBRep are retrained per dataset with shared splits, LoRA settings, checkpoint selection, and evaluation protocols, retaining method-specific mechanisms and initializations (Appendix~\ref{app:baseline_settings}).
With the LoRA-adapted 0.6B text encoder, on Text2CAD, our method achieves Expert R@1=0.640, R@5=0.881, and R@10=0.945, with MedR=1 (Table~\ref{tab:main}). Relative to reproduced CADIR, R@5 and R@10 increase by 24.7 and 22.4 percentage points, respectively, and MedR improves from 2 to 1. The 0.6B LoRA model has the highest reported recall at all three ranks among the systems in Table~\ref{tab:main} on both datasets. The descriptor self-retrieval reference exceeds the text results by 27.3 points at R@1 and 4.4 points at R@10, leaving text--geometry matching as a major source of error.

On MMCAD-B, the same text-backbone configuration achieves R@1=0.371, R@5=0.669, and R@10=0.763, with MedR=2. Relative to reproduced CADIR, R@1 increases by 11.9 percentage points, R@5 by 15.8 points, and R@10 by 13.9 points, while MedR decreases from 5 to 2. The gains cover both first-choice selection and candidate-shortlist retrieval.

Freezing the 0.6B text encoder gives R@1/\allowbreak R@5/\allowbreak R@10 of 0.344/\allowbreak 0.618/\allowbreak 0.731 on Text2CAD and 0.304/\allowbreak 0.622/\allowbreak 0.732 on MMCAD-B, with MedR=3 on both datasets. Without text-encoder adaptation, the model still exceeds CADIR at R@10 on Text2CAD and at all recall ranks on MMCAD-B. 

\subsection{Feature alignment under reordering}
\label{sec:feature_alignment}

\begin{figure}[t]
    \centering
    \includegraphics[width=\columnwidth]{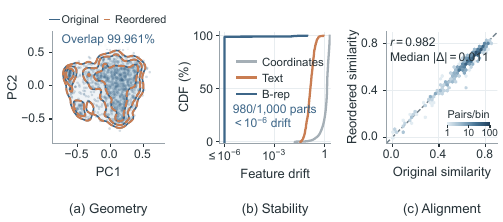}
\caption{\textbf{Feature alignment under reordering on MMCAD-B.} (a) Original and reordered B-rep features in a fixed PCA space; contours enclose 50\%, 80\%, and 95\% probability mass, with overlap computed on a shared density grid. (b) Full-dimensional $\ell_2$ drift of normalized features; Coordinates denotes the sampled input-sequence diagnostic, and values below $10^{-6}$ share a display bin. (c) Original versus reordered text--B-rep matching scores; hexagon color gives pair counts and the dashed line indicates unchanged scores.}
    \label{fig:feature_alignment}
\end{figure}

We evaluate representation stability using the frozen epoch-3 MMCAD-B model (Fig.~\ref{fig:feature_alignment}). For each of the 1,000 test parts, we change contour starting points, traversal directions, and loop order while preserving geometry, yielding 6,514 changed serializations. The projected geometry distributions overlap by 99.961\%; in the full normalized embedding space, 980 parts keep all seven variant distances below $10^{-6}$. For text, reversing or cyclically shifting sentences after the opening sentence produces 746 distinct variants from 373 descriptions, preserving sentence content. Their matched text--B-rep scores correlate with the original scores at $r=0.982$, with median absolute change 0.011. These results support stable geometry representations and preserved cross-modal alignment under the tested reorderings.

\subsection{Ablation studies}
\label{sec:ablation_studies}
\subsubsection{Text adaptation and backbone size}
\label{sec:small_model_tuning}

The main experiments use the 0.6B text backbone. With the Full descriptor fixed, LoRA improves Text2CAD Expert R@5 from 0.618 to 0.881, and MMCAD-B R@5 from 0.622 to 0.669. Appendix~\ref{app:additional_ablations} reports the description-level results and the 4B backbone results.

The 4B LoRA model reaches R@5=0.688 on MMCAD-B and 0.885 on Text2CAD, exceeding the corresponding 0.6B results by 1.9 and 0.4 percentage points, respectively. These results show that the 0.6B backbone achieves recall close to that of the 4B backbone on both datasets.

For the other Text2CAD description levels, the R@5 values are 0.406/0.497 for frozen/LoRA Intermediate descriptions, 0.153/0.180 for Beginner descriptions, and 0.156/0.190 for Abstract descriptions (Table~\ref{tab:small_model_tuning}). Their adaptation gains are 9.1, 2.7, and 3.4 percentage points, respectively, compared with 26.3 points for Expert descriptions. The adaptation benefit therefore varies across description levels, with its largest gain on Expert queries.

\subsubsection{Effect of the number of retained extrusion units}
\label{sec:unit_capacity}
As described in Section~\ref{sec:descriptor}, our descriptor combines a global feature block with local descriptors for up to $U$ extrusion units; the global block still aggregates all units. This ablation varies $U$ to test whether retaining more local geometric information improves retrieval enough to justify a larger descriptor. We compare $U\in\{1,2,4,8\}$ using a LoRA-adapted 0.6B text encoder and Expert descriptions (Table~\ref{tab:unit_capacity}), using the same unit ordering and zero-padding rules (Appendix~\ref{app:retained_units}).

\begin{table}[t]

\centering
\small
\setlength{\tabcolsep}{4.5pt}
\renewcommand{\arraystretch}{1.0}
\setlength{\heavyrulewidth}{0.7pt}
\setlength{\lightrulewidth}{0.35pt}
\setlength{\cmidrulewidth}{0.35pt}
\setlength{\aboverulesep}{0.3ex}
\setlength{\belowrulesep}{0.45ex}
\begin{adjustbox}{max width=\linewidth}
\begin{tabular}{l*{5}{c}}
\toprule
Units $U$ & Dim. $D$ & R@1$\uparrow$ & R@5$\uparrow$ & R@10$\uparrow$ & MedR$\downarrow$ \\
\midrule
1 & 88 & \cellcolor{youtuBlue!4.0}0.284 & \cellcolor{youtuBlue!4.0}0.567 & \cellcolor{youtuBlue!4.0}0.689 & \cellcolor{youtuBlue!4.0}4 \\
2 & 152 & \cellcolor{youtuBlue!9.0}0.378 & \cellcolor{youtuBlue!11.7}0.694 & \cellcolor{youtuBlue!9.4}0.762 & \cellcolor{youtuBlue!17.3}2 \\
4 & 280 & \cellcolor{youtuBlue!23.0}0.640 & \cellcolor{youtuBlue!23.1}0.881 & \cellcolor{youtuBlue!22.8}0.945 & \cellcolor{youtuBlue!24.0}\textbf{1} \\
8 & 536 & \cellcolor{youtuBlue!24.0}\textbf{0.659} & \cellcolor{youtuBlue!24.0}\textbf{0.895} & \cellcolor{youtuBlue!24.0}\textbf{0.961} & \cellcolor{youtuBlue!24.0}\textbf{1} \\
\bottomrule

\end{tabular}
\end{adjustbox}
\caption{\textbf{Effect of the number of retained extrusion units.} LoRA 0.6B, Expert descriptions (1,000 trials; pool=1,000). Bold marks the best retrieval scores. \emph{Color scale:} lighter to darker blue denotes worse to better performance (higher recall or lower median rank), with a separate scale for each metric.}
\label{tab:unit_capacity}
\end{table}
R@5 improves from 0.567 at $U=1$ to 0.694 at $U=2$, 0.881 at $U=4$, and 0.895 at $U=8$. Increasing $U$ from one to four also raises R@1 from 0.284 to 0.640 and reduces MedR from 4 to 1.

$U=8$ achieves the highest recall, exceeding $U=4$ by 1.9, 1.4, and 1.6 percentage points at R@1, R@5, and R@10, respectively; both have MedR=1. The four-unit descriptor uses 280 coordinates, 47.8\% fewer than the 536 coordinates for eight units. We therefore retain $U=4$ as the default trade-off between retrieval performance and descriptor size.
\subsubsection{Contributions of descriptor components}
\label{sec:geometric_information}

With $U=4$, Table~\ref{tab:geometric_ablation} compares descriptor components using LoRA-adapted 4B and 0.6B text encoders. Within each backbone, the Full descriptor and all component-removal variants use the same training and evaluation protocol. Figure~\ref{fig:geometric_ablation} in Appendix~\ref{app:additional_ablations} visualizes the Expert R@1, R@5, and R@10 changes.

\begin{table}[t]

\centering
\small
\setlength{\tabcolsep}{3pt}
\renewcommand{\arraystretch}{1.0}
\setlength{\heavyrulewidth}{0.7pt}
\setlength{\lightrulewidth}{0.35pt}
\setlength{\cmidrulewidth}{0.35pt}
\setlength{\aboverulesep}{0.3ex}
\setlength{\belowrulesep}{0.45ex}
\begin{adjustbox}{max width=\linewidth}
\begin{tabular}{@{}l*{5}{c}@{}}
\toprule
Descriptor variant & $D$ & R@1$\uparrow$ & R@5$\uparrow$ & R@10$\uparrow$ & MedR$\downarrow$ \\
\midrule
\multicolumn{6}{l}{\textit{LoRA Qwen3-Embedding-4B}} \\[2pt]
Full descriptor & 280 & \cellcolor{youtuBlue!24.0}\textbf{0.655} & \cellcolor{youtuBlue!24.0}\textbf{0.885} & \cellcolor{youtuBlue!24.0}\textbf{0.956} & \cellcolor{youtuBlue!24.0}\textbf{1} \\
w/o global block & 256 & \cellcolor{youtuBlue!4.9}0.352 & \cellcolor{youtuBlue!5.9}0.653 & \cellcolor{youtuBlue!4.7}0.741 & \cellcolor{youtuBlue!4.0}3 \\
w/o 2D spectrum & 232 & \cellcolor{youtuBlue!7.3}0.389 & \cellcolor{youtuBlue!6.8}0.664 & \cellcolor{youtuBlue!7.3}0.770 & \cellcolor{youtuBlue!4.0}3 \\
w/o 3D spectrum & 232 & \cellcolor{youtuBlue!7.7}0.396 & \cellcolor{youtuBlue!8.0}0.680 & \cellcolor{youtuBlue!8.3}0.781 & \cellcolor{youtuBlue!14.0}2 \\
w/o turning spectrum & 216 & \cellcolor{youtuBlue!6.6}0.378 & \cellcolor{youtuBlue!6.3}0.658 & \cellcolor{youtuBlue!6.7}0.763 & \cellcolor{youtuBlue!4.0}3 \\
w/o Local stats & 184 & \cellcolor{youtuBlue!4.0}0.337 & \cellcolor{youtuBlue!4.0}0.628 & \cellcolor{youtuBlue!4.0}0.733 & \cellcolor{youtuBlue!4.0}3 \\
\midrule
\multicolumn{6}{l}{\textit{LoRA Qwen3-Embedding-0.6B}} \\[2pt]
Full descriptor & 280 & \cellcolor{youtuBlue!24.0}\textbf{0.640} & \cellcolor{youtuBlue!24.0}\textbf{0.881} & \cellcolor{youtuBlue!24.0}\textbf{0.945} & \cellcolor{youtuBlue!24.0}\textbf{1} \\
w/o global block & 256 & \cellcolor{youtuBlue!5.1}0.348 & \cellcolor{youtuBlue!5.7}0.647 & \cellcolor{youtuBlue!5.2}0.738 & \cellcolor{youtuBlue!4.0}3 \\
w/o 2D spectrum & 232 & \cellcolor{youtuBlue!7.6}0.386 & \cellcolor{youtuBlue!6.7}0.659 & \cellcolor{youtuBlue!7.7}0.766 & \cellcolor{youtuBlue!4.0}3 \\
w/o 3D spectrum & 232 & \cellcolor{youtuBlue!7.9}0.391 & \cellcolor{youtuBlue!8.0}0.676 & \cellcolor{youtuBlue!8.5}0.775 & \cellcolor{youtuBlue!14.0}2 \\
w/o turning spectrum & 216 & \cellcolor{youtuBlue!6.9}0.376 & \cellcolor{youtuBlue!6.0}0.651 & \cellcolor{youtuBlue!7.0}0.758 & \cellcolor{youtuBlue!4.0}3 \\
w/o Local stats & 184 & \cellcolor{youtuBlue!4.0}0.331 & \cellcolor{youtuBlue!4.0}0.625 & \cellcolor{youtuBlue!4.0}0.725 & \cellcolor{youtuBlue!4.0}3 \\
\bottomrule

\end{tabular}
\end{adjustbox}
\caption{\textbf{Descriptor-component ablations with LoRA-adapted text encoders.} Results use Expert descriptions; $D$ is the retained dimension. Recall is a proportion over 1,000 trials (pool=1,000). Bold marks the best retrieval scores within each backbone. \emph{Color scale:} darker blue means higher recall or lower MedR; scales are separate for each metric and backbone.}
\label{tab:geometric_ablation}
\end{table}
The five component blocks comprise the global block (24D), regional spectra ($4\times12$), extrusion spectra ($4\times12$), boundary turning spectra ($4\times16$), and local topology and face statistics ($4\times24$). Each removal configuration omits the indicated block. Differences below compare each component-removal variant with the Full descriptor for the same backbone.

All five removals reduce R@5 for both LoRA-adapted backbones, with the same ordering of score gaps.
Removing local topology and face statistics gives the largest reductions: 25.7 percentage points at 4B and 25.6 at 0.6B.
The global-block reductions are 23.2 and 23.4 points, respectively.
Turning-spectrum removal reduces R@5 by 22.7 and 23.0 points, regional-spectrum removal by 22.1 and 22.2 points, and extrusion-spectrum removal by 20.5 points for both backbones.
The equally sized regional and extrusion blocks (48D each) produce different score gaps, so retained dimensionality alone does not explain this ranking.
The Full descriptor achieves the highest recall at all three ranks and MedR=1, compared with MedR=2 or 3 after component removal.

%% file: section/conclusion.tex
\FloatBarrier
\section{Conclusion}

We introduced SpectraBRep, a framework combining regional, boundary, and extrusion features into an invariant descriptor before cross-modal alignment. A dual encoder aligns the descriptor with natural language through symmetric contrastive learning. Experiments on Text2CAD and MMCAD-B show that a compact, LoRA-adapted text encoder outperforms the compared systems, supporting spectral geometry for language-guided CAD retrieval.

%% file: section/appendix.tex
\FloatBarrier
\section{From construction records to B-rep geometry}
\label{app:geometry}
\subsection{Geometric basis of the sketch--extrusion representation}
\label{app:extrusion_geometry}
\subsubsection{Product geometry and boundary decomposition}
\label{app:product_geometry}
The models in this work combine planar sketches, straight extrusions, and Boolean operations. A straight extrusion preserves its cross-section along the depth direction, allowing the cross-sectional shape and extrusion depth to be described separately.
Let $\Omega\subset\mathbb R^2$ be the planar region enclosed by a sketch, possibly with holes, and let $h>0$ be the extrusion depth. In a local frame aligned with the sketch plane, the solid interior is
\begin{equation}
P=\Omega\times(0,h).
\label{eq:app_product_structure}
\end{equation}
With $I=[0,h]$, the boundary is the union of the end faces $\overline\Omega\times\partial I$ and lateral faces $\partial\Omega\times I$. Thus, $(\Omega,h)$ determines the extrusion geometry, and $\partial\Omega$ generates its lateral B-rep faces.

\subsubsection{Geometric roles of the descriptor components}
\label{app:component_roles}
The three spectral components capture complementary geometric information. The regional Dirichlet spectrum describes the overall cross-section, the boundary turning spectrum captures corners, arcs, and local direction changes, and the extrusion spectrum combines cross-sectional shape with depth.
Different geometries can share a spectrum~\citep{gordon1992drum}. We therefore assess the combined representation's discriminative capacity through the descriptor-distinctness and self-retrieval experiments in Appendix~\ref{sec:descriptor_validation}.

\subsection{Reconstruction of explicit B-rep geometry}
\label{app:reconstruction}
A construction record $X=(c_1,\ldots,c_N)$ specifies an ordered sequence of geometric operations. The reconstruction operator $\mathcal R$ executes them from an empty solid to obtain explicit B-rep geometry:
\begin{equation}
\mathcal R(X)=(F_N\circ\cdots\circ F_1)(\varnothing),
\label{eq:app_reconstruction}
\end{equation}
where $F_t$ is the operation at step $c_t$. The process reconstructs planar profiles, generates extrusion solids, and combines them through Boolean operations.

\subsubsection{Planar profiles}
\label{app:planar_profiles}
A local origin $\mathbf o$, an orthonormal frame $R=[\mathbf e_1\,\mathbf e_2\,\mathbf e_3]$, and a scale $\lambda>0$ embed sketch coordinates into three-dimensional space:
\begin{equation}
\iota(u,v)=\mathbf o+\lambda(u\mathbf e_1+v\mathbf e_2).
\label{eq:app_embedding}
\end{equation}
Adjacent lines or arcs connect through shared endpoints. A closed contour satisfies $\mathbf q_i=\mathbf p_{i+1}$ with cyclic indexing, so connectivity determines the loop geometry while its recorded starting position can vary.
Lines and circular arcs share the generalized-circle equation
\begin{equation}
a(x^2+y^2)+bx+cy+d=0.
\label{eq:app_cline}
\end{equation}
Non-collinear samples determine a circle, while $a=0$ gives a line. Loop orientation and containment identify the outer boundary and holes. The winding number~\citep{jacobson2013winding} characterizes the enclosed region:
\begin{equation}
w(\mathbf p)=\frac{1}{2\pi}\oint_{\partial\Omega}\frac{(\mathbf x-\mathbf p)\times d\mathbf x}{\|\mathbf x-\mathbf p\|^2}.
\label{eq:app_winding}
\end{equation}
Interior points have nonzero winding number. This representation accommodates multiple nested boundary loops.

\subsubsection{Generation of solids}
\label{app:solid_generation}
The recovered cross-section $\Omega$ and depth $h$ generate the straight-extrusion unit in Eq.~\eqref{eq:app_product_structure}. The spectral derivation uses this product structure to separate cross-sectional and depth variation.

\subsubsection{Boolean composition}
\label{app:boolean_composition}
Units are combined according to the recorded Boolean operations. Let $f$ and $g$ be scalar fields for the existing solid and new feature, with negative values inside. A new body uses $f'=g$; union, intersection, and subtraction use
\begin{equation}
f'=\begin{cases}
\min(f,g), & \text{union},\\
\max(f,g), & \text{intersection},\\
\max(f,-g), & \text{subtraction}.
\end{cases}
\label{eq:app_boolean}
\end{equation}
These updates describe the Boolean composition. Retrieval features are constructed from the recovered B-rep regions, boundaries, extrusion units, and their relations.

\subsection{Equivariance of reconstruction}
\label{app:reconstruction_equivariance}
\subsubsection{Similarity transformations and discretization}
\label{app:similarity_conditions}
Consider a similarity $g(\mathbf x)=aQ\mathbf x+\mathbf t$, where $a>0$ and $Q\in O(3)$. It preserves connectivity, loop topology, extrusion relations, and Boolean composition. Preserving these relations numerically under uniform scaling requires dimensionless thresholds and equivariant sampling rules.
Scaling multiplies curvature $\kappa$ by $a^{-1}$ and chord length $\ell$ by $a$, preserving $|\kappa'|\ell'=|\kappa|\ell$. Boundary sampling likewise uses dimensionless angular increments. Under these conditions, reconstruction commutes with the similarity action:
\begin{equation}
\mathcal R(g\cdot X)=g\cdot\mathcal R(X).
\label{eq:app_reconstruction_equivariance}
\end{equation}
For a descriptor satisfying $\Phi(g\cdot\mathcal B)=\Phi(\mathcal B)$, the composition $\Phi\circ\mathcal R$ is similarity invariant.

\subsubsection{Operation-level equivariance and recording conventions}
\label{app:operation_equivariance_details}
The action on a construction record transforms sketch positions, local frames, and extrusion directions, scales lengths by $a$, and preserves operation types and incidence relations. An operation $F$ and its transformed version $F^g$ satisfy
\begin{equation}
F^g(g\cdot B)=g\cdot F(B).
\label{eq:app_operation_equivariance}
\end{equation}
Sketch embedding and straight extrusion satisfy this relation under the corresponding parameter transformations. Bijectivity gives $g(A\circ B)=g(A)\circ g(B)$ for union, intersection, and subtraction. Starting from $g(\varnothing)=\varnothing$, induction over construction steps yields Eq.~\eqref{eq:app_reconstruction_equivariance}. Its numerical realization uses the threshold and discretization conditions above.
Cyclic contour shifts and traversal reversal preserve connectivity and the enclosed region. Unit reindexing preserves reconstructed geometry when it relabels the same units and Boolean relations while retaining operation dependencies.

\section{Invariance properties of the spectral descriptor}
\label{app:spectral_invariance}
\subsection{Similarity invariance of the Dirichlet spectrum}
\label{app:dirichlet_invariance}
\subsubsection{Eigenfunction transformation and Laplacian scaling}
\label{app:dirichlet_transform}
Let $\Omega\subset\mathbb R^2$ be a bounded Lipschitz domain of area $A>0$, with Dirichlet eigenvalues ordered increasingly with multiplicity. For a similarity $g$ with scale $a>0$ and orthogonal part $Q\in O(2)$, an eigenpair $(\mu,u)$ induces the transformed eigenfunction
\begin{equation}
u'(\mathbf x')=u(g^{-1}(\mathbf x')).
\label{eq:app_transformed_eigenfunction}
\end{equation}
This function vanishes on the transformed boundary. Orthogonality of $Q$ and the chain rule give the following scaling law, which also holds in the weak formulation:
\begin{equation}
-\Delta u'=a^{-2}\mu u'.
\label{eq:app_laplacian_scaling}
\end{equation}
\subsubsection{Scale normalization and contour conventions}
\label{app:dirichlet_normalization}
The transformation is invertible and preserves eigenspace multiplicities, so $\mu_k'=a^{-2}\mu_k$. Since $A'=a^2A$, multiplication cancels the scale factor:
\begin{equation}
\mu_k'A'=\mu_k A.
\label{eq:app_dirichlet_invariance}
\end{equation}
Thus, $\widehat\mu_k=\mu_k|\Omega|$ is invariant to translation, rotation, reflection, and uniform scaling. Contour-origin and traversal changes leave the region and its eigenvalue problem unchanged. The same result holds for cross-sections with holes when Dirichlet conditions are imposed on every boundary component.

\subsection{Extrusion spectrum by separation of variables}
\label{app:extrusion_spectrum}
\subsubsection{Product modes and eigenvalues}
\label{app:extrusion_modes}
Impose Dirichlet conditions on the end and lateral faces of $P=\Omega\times(0,h)$. The Laplacian separates as $\Delta_P=\Delta_\Omega+\partial_{zz}$. For a cross-sectional eigenfunction $u_m$ and positive integers $m,n$, take the product mode
\begin{equation}
w(\mathbf x,z)=u_m(\mathbf x)\sin(n\pi z/h).
\label{eq:app_product_mode}
\end{equation}
The cross-sectional factor vanishes on the lateral faces, and the sine factor vanishes on both end faces. Differentiation gives
\begin{equation}
-\Delta w=\bigl[\mu_m+(n\pi/h)^2\bigr]w.
\label{eq:separation}
\end{equation}
The products of the two eigenfunction bases form a complete basis on the product domain, with eigenvalues
\begin{equation}
\lambda_{m,n}=\mu_m+(n\pi/h)^2.
\label{eq:app_extrusion_spectrum}
\end{equation}
Sorting with multiplicity gives the lowest 12 values used by the descriptor. More generally, $1\leq m,n\leq K$ covers the lowest $K$ values: every candidate outside this range has at least $K$ candidates no larger than itself. The model-level descriptor assembles these unit spectra together with composition attributes.
\subsubsection{Scale normalization}
\label{app:extrusion_normalization}
Write $V=Ah$ and $\widehat\lambda=\lambda V^{2/3}$. Uniform scaling multiplies eigenvalues by $a^{-2}$ and volume by $a^3$, yielding
\begin{equation}
\widehat\lambda'=\widehat\lambda.
\label{eq:app_extrusion_invariance}
\end{equation}
\subsubsection{Relative extrusion depth}
\label{app:relative_depth}
Using the dimensionless relative depth $r=h/\sqrt A$ and area-normalized regional eigenvalues gives
\begin{equation}
\widehat\lambda_{m,n}=\widehat\mu_m r^{2/3}+(n\pi)^2r^{-4/3}.
\label{eq:app_relative_extrusion}
\end{equation}
This expression combines cross-sectional shape with relative depth while removing absolute size.

\subsection{Invariance of the boundary turning Fourier spectrum}
\label{app:turning_invariance}
\subsubsection{Continuous turning measure}
\label{app:turning_measure}
Consider a piecewise smooth closed contour of perimeter $L$, parameterized by arc length $s$. The signed turning measure $d\theta$, measured along the chosen traversal direction, has Fourier coefficients
\begin{equation}
\mathcal F_k=\int_0^L e^{-2\pi\mathrm i ks/L}\,d\theta(s).
\label{eq:app_continuous_turning}
\end{equation}
Smooth segments contribute $\kappa(s)\,ds$, while a corner at $s_v$ contributes a point mass of signed turn $\alpha_v$. Straight segments have zero curvature, circular arcs accumulate turning along their length, and corners contribute discrete turns.
\subsubsection{Discrete coefficients and sampling conditions}
\label{app:turning_sampling}
With turning increments $\Delta\theta_j$ recorded at arc-length positions $s_j$, the discrete coefficient at integer frequency $k$ is
\begin{equation}
F_k=\sum_j\Delta\theta_j e^{-2\pi\mathrm i ks_j/L}.
\label{eq:app_turning_fourier}
\end{equation}
For polygons, corner turns represent the turning measure exactly. Curved segments use accumulated turning over sampled intervals or a geometrically equivariant polygonal approximation. The transformation laws below require corresponding sampling positions and turning increments; dimensionless sampling rules make this correspondence compatible with uniform scaling.
\subsubsection{Change of contour origin}
\label{app:turning_origin}
Shifting the origin by $s_0$ maps positions to $s_j'=(s_j-s_0)\bmod L$. At integer frequencies,
\begin{equation}
F_k'=e^{2\pi\mathrm i ks_0/L}F_k.
\label{eq:app_turning_origin}
\end{equation}
The phase factor has unit modulus, so the magnitude is unchanged.
\subsubsection{Reversal of traversal}
\label{app:turning_reversal}
With the origin fixed, traversal reversal gives $s_j'=(-s_j)\bmod L$ and $\Delta\theta_j'=-\Delta\theta_j$. Since the increments are real,
\begin{equation}
F_k'=-\overline{F_k}.
\label{eq:app_turning_reversal}
\end{equation}
An additional origin shift multiplies this expression by the unit-modulus factor in Eq.~\eqref{eq:app_turning_origin}.
\subsubsection{Spatial transformations}
\label{app:turning_spatial}
Translation and rotation preserve arc length and signed turning, giving $F_k'=F_k$. Reflection reverses signed turning when the mapped traversal order is retained, giving $F_k'=-F_k$. Uniform scaling changes arc length and perimeter equally, preserving turning angles and normalized positions. Each transformation and their compositions therefore satisfy
\begin{equation}
|F_k'|=|F_k|.
\label{eq:app_turning_invariance}
\end{equation}
The continuous coefficients $\mathcal F_k$ obey the same laws. For a cross-section with several contours, the result applies to each component, with their geometric relationships retained during assembly.

\subsection{Rotational symmetry estimation from harmonic concentration}
\label{app:rotational_symmetry}
\label{sec:symmetry}
\subsubsection{Harmonic structure of rotational symmetry}
\label{app:symmetry_harmonics}
For a contour with $q$-fold rotational symmetry, the turning pattern repeats every $L/q$ along arc length. If $G_k$ is the contribution from one repeated segment, then
\begin{equation}
F_k=G_k\sum_{\ell=0}^{q-1}e^{-2\pi\mathrm i k\ell/q}.
\label{eq:app_symmetry_harmonics}
\end{equation}
The geometric sum is $q$ when $q$ divides $k$ and zero otherwise. Nonzero harmonics of an ideally symmetric contour therefore occur at multiples of $q$. This holds exactly for the continuous measure and for discrete sampling that preserves the repeated structure.
\subsubsection{Finite-band symmetry estimation}
\label{app:symmetry_estimation}
The evaluated positive frequencies form a finite set $\mathcal K$, with total energy $E=\sum_{k\in\mathcal K}|F_k|^2$. When $E>0$, the energy concentration at order $q$ is
\begin{equation}
\rho(q)=\frac{\sum_{k\in\mathcal K,\,q\mid k}|F_k|^2}{E}.
\label{eq:symmetry_ratio}
\end{equation}
From a finite candidate set $\mathcal Q$ containing $q=1$, select the largest order with concentration at least 0.98:
\begin{equation}
\widehat q=\max\{q\in\mathcal Q:\rho(q)\geq0.98\}.
\label{eq:app_symmetry_estimate}
\end{equation}
The candidate $q=1$ ensures a solution for positive energy. We exclude $F_0$, which records the total turn $\pm2\pi$ of a simple closed contour and belongs to the multiples of every integer. The estimated order $\widehat q$ depends on the chosen frequency band and threshold.
\subsubsection{Degenerate cases and geometric consistency}
\label{app:symmetry_degenerate}
Zero or numerically unresolved positive-frequency energy requires analytic contour geometry or a wider band to determine symmetry. An ideal circle has zero coefficients at every nonzero frequency; symmetry above the evaluated band can also yield zero energy. For a cross-section with holes, the same rotation must preserve all boundary components and their relative positions. The resulting contour-symmetry attribute enters the geometric descriptor.

\section{B-rep descriptor layout}
\label{app:descriptor_layout}
\begin{figure}[t]
\centering
\includegraphics[width=\textwidth]{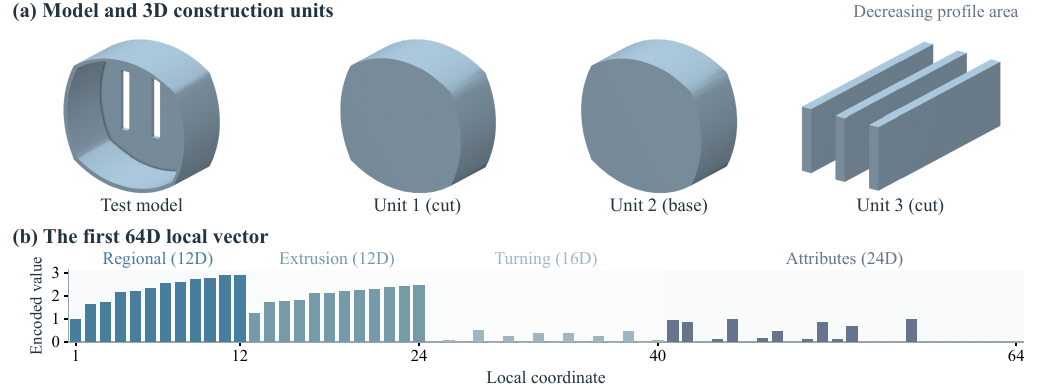}
\caption{\textbf{Assembly of the 280D descriptor from a real Text2CAD test model.} (a) The reconstructed solid (\texttt{0000/00007648}) and its recorded unit profiles, stably ordered by decreasing area. (b) Actual coordinates of the first 64D local vector: 12 regional, 12 extrusion, 16 turning, and 24 geometric attributes.}
\label{fig:descriptor_assembly}
\end{figure}

The experimental descriptor is stored in \texttt{float32} as a global block followed by four local slots:
\begin{equation}
\mathbf d=[\mathbf g;\mathbf v_1;\mathbf v_2;\mathbf v_3;\mathbf v_4].
\label{eq:app_descriptor_dimension}
\end{equation}
The global block has 24 coordinates and each local slot has 64, giving 280 in total. A local vector contains 12 regional spectral features, 12 extrusion spectral features, 16 turning Fourier magnitudes, and 24 geometric and record-level attributes.
The following subsections describe unit extraction, feature computation, model-level assembly, and input preprocessing. Tables~\ref{tab:descriptor_layout} and~\ref{tab:descriptor_global_features} list all coordinates before standardization, and Figure~\ref{fig:descriptor_assembly} illustrates assembly for a real test model.

\subsection{Unit extraction and numerical features}
\label{app:unit_features}

\paragraph{Source of extrusion units.}
The constructor reads \texttt{cad\_vec} and its unit indices \texttt{index\_vec}. Each unit is defined by recorded sketch faces, boundary curves, two extrusion extents, and an operation code. This unit decomposition follows the construction history rather than the faces of the final Boolean solid.

The first-channel tokens delimit the record: 0 is padding and 1 is skipped; 5, 4, and 3 end a curve, loop, and sketch face, respectively. Token 2 starts the extrusion parameters and supplies the unit index, while token 6 commits the unit. Tokens 7--10 encode operation codes 0--3.

For a coordinate row $(a,b)$ with $a\geq11$, sketch coordinates are $(a-11,b-11)$; scalars in the extrusion block are $(a-11)/255$. The first two scalars give $e_1,e_2$, with missing values set to zero and depth $h=e_1+e_2$. Units are initially returned in increasing index order, retaining the last committed record for a repeated index.

The descriptor uses sketch coordinates, extrusion extents, and operation codes. Recorded offsets, Euler angles, and sketch-scale fields are excluded from feature computation. STEP inputs therefore require conversion to these construction records.

\paragraph{Polygonal boundaries and representative face.}
A line connects its recorded start to the next curve's start. An arc is reconstructed from its start, intermediate point, and the next start, then sampled at 12 equally spaced angles including both endpoints. The final endpoint is omitted during concatenation. A collinearity determinant below $10^{-9}$ instead gives 12 linear samples.

A loop consisting of one two-point curve denotes a circle: the first point is its center, the second determines its radius, and 48 samples are placed at $2\pi r/48$ for $r=0,\ldots,47$. Curve counts and type fractions use the recorded primitives.

Loops with fewer than three polygon vertices are discarded. In each sketch face, the loop with the largest absolute shoelace area is the outer loop, and the other valid loops are holes. The unit's representative face has the largest outer-loop area. Both selections retain the first occurrence in an exact area tie; units without a valid face are discarded.
For a loop with vertices $p_r$, the area is
\begin{equation}
A=\frac12\left|\sum_r p_r\times p_{r+1}\right|.
\label{eq:app_loop_attributes}
\end{equation}
Its perimeter is $L=\sum_r\|p_{r+1}-p_r\|$, and its isoperimetric ratio is $I=\operatorname{clip}(4\pi A/L^2,0,1.05)$, with $I=0$ when $L\leq10^{-9}$. The aspect ratio $\rho$ is the square root of the smaller-to-larger eigenvalue ratio of the vertices' population covariance matrix, with each eigenvalue floored at $10^{-12}$.

For unit $j$, write the representative outer-loop attributes as $A_j,I_j,\rho_j$ and its face's hole set as $H_j$. The counts $f_j$ and $b_j$ are the number of valid sketch faces and their total holes, respectively, at the construction-record level. Relative depth is $\ell_j=h_j/\max(\sqrt{A_j}/255,10^{-6})$, and extrusion asymmetry is $\eta_j=|e_{1j}-e_{2j}|/\max(h_j,10^{-9})$.

\subsubsection{Spectral features}
\label{app:unit_spectra}
The following formulas describe one unit, omitting its index; the feature tables restore unit indices.

\paragraph{Regional block.}
Each face is rasterized on a $48\times48$ grid. Let $c$ be the center of the outer loop's axis-aligned bounding box and $a$ its largest side length. Boundary points map to $(p-c)45/a+(24,24)$; a span below $10^{-9}$ gives an empty mask.
The horizontal ray-crossing test marks pixel centers $(x+0.5,y+0.5)$ inside the outer polygon and outside all holes as occupied. An edge straddles the ray when $y_0\leq y<y_1$ or $y_1\leq y<y_0$; only intersections strictly to the right of the test point are counted.

On the representative face's $n$ occupied cells, the five-point Dirichlet matrix $D$ has diagonal 4, off-diagonal $-1$ for occupied four-neighbors, and zero boundary values outside the mask. Its 12 smallest eigenvalues $\gamma_k$ give the regional inputs
\begin{equation}
\beta_k=\operatorname{clip}(n\gamma_k,0,10^6),\quad k=1,\ldots,12.
\label{eq:app_raster_spectrum}
\end{equation}
The occupied-cell count approximates area normalization. We use \texttt{eigsh} with $k=12$, \texttt{sigma}$=-10^{-6}$, \texttt{which=LM}, and \texttt{maxiter=5000}, then sort the results increasingly. Fewer than 16 occupied cells or a solver exception yields twelve zeros. The face fill ratio is $r=n/48^2$.

\paragraph{Normalized extrusion spectrum.}
The extrusion vector $\mathbf q$ contains the twelve smallest normalized three-dimensional Dirichlet eigenvalues. Equation~\eqref{eq:app_relative_extrusion} computes them from the regional spectrum and relative depth, with area and depth expressed in consistent geometric units. This implements the normalized spectrum in Eq.~\eqref{eq:normalized_extrusion}. All twelve entries are zero when $h\leq10^{-6}$ or every $\beta_m$ is zero.
Both spectral blocks use the coordinatewise logarithmic transform
\begin{equation}
\psi(x)=\log\left(1+\frac{\max(x,0)}{10}\right).
\label{eq:app_spectral_transform}
\end{equation}

\paragraph{Turning block and symmetry code.}
Discard edges of length at most $10^{-9}$ from the representative outer polygon. For the remaining lengths $l_r$ and directions $\theta_r$, compute $\Delta\theta_r=\operatorname{wrap}(\theta_r-\theta_{r-1})\in[-\pi,\pi)$ and normalized arc-length positions $u_r=L^{-1}\sum_{t<r}l_t$.
Evaluate Eq.~\eqref{eq:app_turning_fourier} at $s_r/L=u_r$ and take magnitudes $M_k=|F_k|$. The first sixteen form $\mathbf M$, giving the turning block $\mathbf b=\mathbf M/\|\mathbf M\|_2$. The block is zero when its norm is at most $10^{-9}$, fewer than three edges remain, or the perimeter is below $10^{-9}$.

Symmetry estimation uses the first 24 harmonics and candidate orders 1--12. Let $E$ be the energy in Appendix~\ref{app:symmetry_estimation} and $T=|\sum_r\Delta\theta_r|$ the absolute total turn. Energy below $(0.02T)^2$ gives code 0. Otherwise, select the largest order reaching concentration 0.98 in Eq.~\eqref{eq:symmetry_ratio}, defaulting to 1. An undefined turning measure also gives code 1. We denote unit $j$'s code by $s_j$, with 0 marking the low-energy case.

\subsubsection{Local topology and geometric statistics}
\label{app:unit_statistics}
The tables use the clipped count $c_t(x)=\min(x,t)/t$, reciprocal symmetry feature $R(s)=1/\max(s,1)$, and denominator floor $\epsilon=10^{-9}$. A bar and $\operatorname{sd}$ denote the mean and population standard deviation.

The indicator $C_j$ marks a recorded circular outer loop. The counts $N_j$, $N_j^{\rm arc}$, and $N_j^{\rm line}$ refer to all primitives, arcs, and lines, respectively. A circle counts as one primitive and zero arcs or lines. The operation code $o_j$ defaults to $-1$; its local one-hot feature uses $o_j^+=\min(\max(o_j,0),3)$.
Hole attributes $A_a,I_a,\rho_a,C_a$ use the same definitions. When $H_j$ is empty, all six hole-summary coordinates (59--64) are zero.

\begin{table}[t]

\centering
\small
\setlength{\tabcolsep}{7pt}
\renewcommand{\arraystretch}{1.18}
\begin{adjustbox}{max width=\linewidth}
\begin{tabular}{c l c l}
\toprule
\rowcolor{black!5}
\makebox[3.4em][c]{\textbf{Index}} & \textbf{Feature} & \textbf{Dim.} & \textbf{Value before standardization} \\
\midrule
\textcolor{black!65}{1--12} & Regional spectrum & 12 & $\psi(\beta_{j,1:12})$ \\
\textcolor{black!65}{13--24} & Depth-combined spectrum & 12 & $\psi(q_{j,1:12})$ \\
\textcolor{black!65}{25--40} & Outer turning spectrum & 16 & $M_{1:16}/\|M_{1:16}\|_2$ (zero rule in text) \\
\addlinespace[3pt]
\textcolor{black!65}{41} & Outer isoperimetric ratio & 1 & $I_j$ \\
\textcolor{black!65}{42} & Outer aspect ratio & 1 & $\rho_j$ \\
\textcolor{black!65}{43} & Outer circle indicator & 1 & $C_j$ \\
\textcolor{black!65}{44} & Outer primitive count & 1 & $c_{32}(N_j)$ \\
\textcolor{black!65}{45} & Outer arc fraction & 1 & $N_j^{\rm arc}/\max(N_j,1)$ \\
\textcolor{black!65}{46} & Outer line fraction & 1 & $N_j^{\rm line}/\max(N_j,1)$ \\
\textcolor{black!65}{47} & Symmetry code & 1 & $c_{12}(s_j)$ \\
\textcolor{black!65}{48} & Reciprocal symmetry code & 1 & $R(s_j)$ \\
\addlinespace[3pt]
\textcolor{black!65}{49} & Unit sketch-hole count & 1 & $c_8(b_j)$ \\
\textcolor{black!65}{50} & Unit sketch-face count & 1 & $c_8(f_j)$ \\
\textcolor{black!65}{51} & Relative depth & 1 & $\tanh(\ell_j)$ \\
\textcolor{black!65}{52} & Extrusion asymmetry & 1 & $\eta_j$ \\
\textcolor{black!65}{53} & Representative raster fill & 1 & $r_j$ \\
\textcolor{black!65}{54} & Representative hole-area fraction & 1 & $\min(\sum_{a\in H_j} A_a/\max(A_j,\epsilon),1)$ \\
\textcolor{black!65}{55--58} & Operation one-hot & 4 & $\mathbf1[o_j^+=t]$, $t=0,1,2,3$ \\
\addlinespace[3pt]
\textcolor{black!65}{59} & Mean hole isoperimetric ratio & 1 & $\operatorname{mean}_{a\in H_j} I_a$ \\
\textcolor{black!65}{60} & Spread of hole isoperimetric ratios & 1 & $\operatorname{sd}_{a\in H_j} I_a$ \\
\textcolor{black!65}{61} & Circular-hole fraction & 1 & $\operatorname{mean}_{a\in H_j} C_a$ \\
\textcolor{black!65}{62} & Relative spread of hole areas & 1 & $\operatorname{sd}_{a\in H_j} A_a/\max(\operatorname{mean}_{a\in H_j} A_a,\epsilon)$ \\
\textcolor{black!65}{63} & Representative hole count & 1 & $c_8(|H_j|)$ \\
\textcolor{black!65}{64} & Mean hole aspect ratio & 1 & $\operatorname{mean}_{a\in H_j} \rho_a$ \\
\bottomrule
\end{tabular}
\end{adjustbox}
\caption{\textbf{Exact 64-dimensional local layout.} Indices are one-based within each unit slot. The first 40 coordinates are spectral features; coordinates 41--64 are the 24 additional attributes.}
\label{tab:descriptor_layout}
\end{table}

\Needspace{8\baselineskip}
\subsection{Model-level aggregation and layout}
\label{app:model_aggregation}

\subsubsection{Geometric ordering of extrusion units}
\label{app:unit_ordering}
Let $m$ be the number of valid units and $\mathbf u_j$ the local descriptor of unit $j$. Units are stably sorted by decreasing representative outer-loop area $A_j$, with no secondary key. Exact area ties retain the parser's unit-index order. Let $\sigma$ denote the resulting permutation. For occupied slots, with $r$ from 1 to $\min(m,4)$, we set
\begin{equation}
\mathbf v_r=\mathbf u_{\sigma(r)}.
\label{eq:app_model_assembly}
\end{equation}
The remaining slots up to four are zero vectors. Raw coordinates 1--24 contain $\mathbf g$, and coordinates $25+64(r-1)$ through $88+64(r-1)$ contain slot $r$. Padding is applied before standardization; a raw zero slot generally becomes nonzero after standardization.

\subsubsection{Global aggregation and descriptor dimension}
\label{app:global_aggregation}
The global block aggregates all $m$ valid units, including the four with local slots and every additional unit. In Table~\ref{tab:descriptor_global_features}, $B=\sum_j b_j$ is the total hole count, $F=\sum_j f_j$ the total sketch-face count, and $S=\max_j s_j$ the maximum symmetry code.
The counts $N_{\rm all}$ and $C_{\rm all}$ cover primitives and recorded circles across all loops of all valid sketch faces. Global operation counts use raw codes, so a missing code $-1$ contributes to none of coordinates 8--11; the local one-hot feature uses the clamped code. All means and standard deviations include every valid unit.

\begin{table}[t]

\centering
\small
\setlength{\tabcolsep}{7pt}
\renewcommand{\arraystretch}{1.18}
\begin{adjustbox}{max width=\linewidth}
\begin{tabular}{c l c l}
\toprule
\rowcolor{black!5}
\makebox[3.4em][c]{\textbf{Index}} & \textbf{Feature} & \textbf{Dim.} & \textbf{Value before standardization} \\
\midrule
\textcolor{black!65}{1} & Valid-unit count & 1 & $c_8(m)$ \\
\textcolor{black!65}{2} & Total sketch-hole count & 1 & $c_{12}(B)$ \\
\textcolor{black!65}{3} & Total primitive count & 1 & $c_{64}(N_{\rm all})$ \\
\textcolor{black!65}{4} & Total circular-loop count & 1 & $c_{12}(C_{\rm all})$ \\
\addlinespace[3pt]
\textcolor{black!65}{5} & Maximum numerical symmetry code & 1 & $c_{12}(S)$ \\
\textcolor{black!65}{6} & Reciprocal maximum code & 1 & $R(S)$ \\
\textcolor{black!65}{7} & Zero maximum-code indicator & 1 & $\mathbf1[S=0]$ \\
\textcolor{black!65}{8--11} & Operation counts & 4 & $c_4(\sum_j\mathbf1[o_j=t])$, $t=0,1,2,3$ \\
\addlinespace[3pt]
\textcolor{black!65}{12} & Mean relative depth & 1 & $\tanh(\operatorname{mean}_j\ell_j)$ \\
\textcolor{black!65}{13} & Spread of relative depths & 1 & $\tanh(\operatorname{sd}_j\ell_j)$ \\
\textcolor{black!65}{14} & Mean outer isoperimetric ratio & 1 & $\operatorname{mean}_j I_j$ \\
\textcolor{black!65}{15} & Spread of outer isoperimetric ratios & 1 & $\operatorname{sd}_j I_j$ \\
\textcolor{black!65}{16} & Smallest-to-largest outer area & 1 & $\min_j A_j/\max(\max_j A_j,\epsilon)$ \\
\textcolor{black!65}{17} & Relative depth spread & 1 & $\operatorname{sd}_j h_j/\max(\operatorname{mean}_j h_j,\epsilon)$ \\
\textcolor{black!65}{18} & Mean outer aspect ratio & 1 & $\operatorname{mean}_j\rho_j$ \\
\textcolor{black!65}{19} & Mean representative raster fill & 1 & $\operatorname{mean}_j r_j$ \\
\textcolor{black!65}{20} & Total sketch-face count & 1 & $c_{12}(F)$ \\
\addlinespace[3pt]
\textcolor{black!65}{21--24} & Mean transformed depth spectrum & 4 & $m^{-1}\sum_j\psi(q_{j,k})$, $k=1,2,3,4$ \\
\bottomrule
\end{tabular}
\end{adjustbox}
\caption{\textbf{Exact 24-dimensional global layout.} This block occupies coordinates 1--24 of the raw model descriptor and uses all valid units.}
\label{tab:descriptor_global_features}
\end{table}

\subsection{Assembly procedure and standardization}
\label{app:assembly_conditions}

\subsubsection{Retained units, padding, and overflow}
\label{app:retained_units}
The complete construction follows these steps:
\begin{enumerate}
\item Parse the construction records by unit index. Reconstruct polygonal loops, discard invalid loops and faces, and retain units with at least one valid face.
\item For each valid face, identify its outer loop and holes and compute their polygon statistics and raster spectrum. Select the representative face by outer-loop area.
\item For every unit, compute $h_j,\ell_j,\eta_j$, its depth-combined spectrum, outer turning harmonics and symmetry code, sketch-face and hole counts, and its 64-dimensional vector in Table~\ref{tab:descriptor_layout}.
\item Stably sort units by decreasing $A_j$. Compute the global vector from all units using Table~\ref{tab:descriptor_global_features}.
\item Initialize $\mathbf d=\mathbf0_{280}$ in \texttt{float32}; write $\mathbf g$ into the first 24 coordinates, then write the first $\min(m,4)$ unit vectors into successive 64-dimensional slots. Leave missing slots at zero. Additional units contribute through the global block.
\end{enumerate}
If no valid unit remains, the vector routine returns $\mathbf0_{280}$. The data-cache builder excludes such records, as well as missing inputs and records that raise parsing exceptions, from the retained model set. Raster-solver failures alone retain the unit with the zero spectral block described above.

After the feature-specific transforms, fit a mean and sample standard deviation for each coordinate using training models. For training values $x_1,\ldots,x_n$, compute $\mu=n^{-1}\sum_i x_i$ and $s^2=(n-1)^{-1}\sum_i(x_i-\mu)^2$, then standardize by
\begin{equation}
\widetilde x=\frac{x-\mu}{\max(s,10^{-4})}.
\label{eq:app_descriptor_standardization}
\end{equation}
Evaluation reuses the training values of $\mu$ and $s$. Construction proceeds through \texttt{parse\_geometry}, \texttt{describe}, and \texttt{descriptor\_vector}, followed by standardization. Rounding each standardized coordinate to the nearest integer produces the 280-dimensional quantized key supplied to the geometry encoder.

\subsubsection{Invariance conditions of the implemented assembly}
\label{app:invariant_assembly}
The raster spectrum and sampled turning harmonics provide numerical approximations. The extrusion block uses the normalized three-dimensional spectrum in Eq.~\eqref{eq:app_relative_extrusion}. Stable sorting by decreasing area removes unit-index dependence when areas are distinct. Exact area ties preserve the input unit-index order.

\section{Descriptor validation}
\label{app:validation_details}
\label{sec:descriptor_validation}
\paragraph{Numerical consistency and invariance.}
In the analytic-shape checks, cyclic contour shifts and traversal reversal give zero error, and turning-spectrum errors remain below $3\times10^{-15}$. The finite-resolution planar rotation-and-scaling check has a maximum relative error of 0.015. Analytic-shape checks and numerical details appear in Table~\ref{tab:validation}.
\paragraph{Discriminativeness and self-retrieval.}
Among 4,000 randomly sampled test models, the descriptor yields 3,315 distinct representations, compared with 3,601 distinct raw construction sequences, a count ratio of about 92\%. Querying with the target model's own descriptor gives R@1=0.913, R@5=0.951, and R@10=0.989. This geometry-only self-retrieval result provides a reference for interpreting text-to-CAD retrieval performance. We also examine distances within duplicate models and between different models below.

\subsection{Numerical validation details}
Table~\ref{tab:validation} summarizes checks of spectral accuracy, invariance, and rotational symmetry. For the unit square, the reference eigenvalues are $\pi^2(m^2+n^2)$ in the limit of zero grid spacing; the observed bias is uniform. For the unit disk, the normalized first eigenvalue $\mu_1|\Omega|$ is 17.55, compared with the analytic value $j_{01}^2\pi=18.17$.
The planar transformation combines a $37^\circ$ rotation with a scale factor of 0.4. Contour reindexing includes cyclic shifts and traversal reversal; the turning spectrum is evaluated under all tested transformations. The theoretical invariance error is zero. Symmetry order is evaluated on regular polygons with 3--12 sides.
\begin{table}[t]

\centering
\small
\setlength{\tabcolsep}{4.5pt}
\renewcommand{\arraystretch}{1.0}
\setlength{\heavyrulewidth}{0.7pt}
\setlength{\lightrulewidth}{0.35pt}
\setlength{\cmidrulewidth}{0.35pt}
\setlength{\aboverulesep}{0.3ex}
\setlength{\belowrulesep}{0.45ex}
\begin{adjustbox}{max width=\linewidth}
\begin{tabular*}{\columnwidth}{@{\extracolsep{\fill}}ll@{}}
\toprule
Check & Result \\
\midrule
Square eigenvalues & $-4.4\%$ bias \\
Disk eigenvalue & $-3.4\%$ error \\
\addlinespace[2pt]
Rotation and scaling & $1.5\%$ max. error \\
Contour reindexing & Zero error \\
Turning spectrum & $\leq 3\times10^{-15}$ error \\
\addlinespace[2pt]
Symmetry order & $10/10$ correct \\
\bottomrule
\end{tabular*}
\end{adjustbox}
\caption{\textbf{Numerical validation of the spectral descriptor.}}
\label{tab:validation}
\end{table}
\subsection{Complete-descriptor transformation study}
\label{app:full_descriptor_stability}
We randomly sample 1,000 Text2CAD test models and evaluate global translation, rotation, reflection, and three contour-recording changes: cyclic starting-point shift, traversal reversal, and their combination. Of these models, 988 reconstruct successfully and 987 pass the B-rep consistency checks. Global transformations preserve the descriptor's unit-local inputs. Contour transformations modify the vertex records before descriptor recomputation; excluding identity cases gives 809 nontrivial pairs per contour test.

For source descriptor $\mathbf d$ and transformed-record descriptor $\mathbf d'$, we measure the complete 280D relative difference
\begin{equation}
E_{\mathrm{full}}=
\frac{\|\mathbf d'-\mathbf d\|_2}
{\max(\|\mathbf d\|_2,10^{-9})}.
\label{eq:full_descriptor_stability_error}
\end{equation}
Table~\ref{tab:full_descriptor_stability} reports the fraction within 1\% relative error, the maximum error, and quantized-key agreement. Keys are computed using the input preprocessing in Section~\ref{app:assembly_conditions}; agreement requires equality in all 280 coordinates. This evaluates invariance and cross-record collisions at the geometry encoder's input.

\begin{table}[t]

\centering
\small
\setlength{\tabcolsep}{5pt}
\renewcommand{\arraystretch}{1.12}
\begin{adjustbox}{max width=\linewidth}
\begin{tabular}{@{}lccccc@{}}
\toprule
\multicolumn{6}{l}{\textbf{(a) Complete 280D stability under the six tested transformations}}\\[2pt]
Transformation & Certified & Compared & $E_{\mathrm{full}}\!\leq\!1\%$ (\%) & Key agreement (\%) & Max. $E_{\mathrm{full}}$ (\%)\\
\midrule
Global translation & 987 & 987 & \cellcolor{youtuBlue!10}100.00 & \cellcolor{youtuBlue!10}100.00 & 0 \\
Global $37^\circ$ rotation & 987 & 987 & \cellcolor{youtuBlue!10}100.00 & \cellcolor{youtuBlue!10}100.00 & 0 \\
Global reflection & 987 & 987 & \cellcolor{youtuBlue!10}100.00 & \cellcolor{youtuBlue!10}100.00 & 0 \\
Contour starting-point shift & 987 & 809 & \cellcolor{youtuBlue!10}99.88 & \cellcolor{youtuBlue!10}99.75 & 9.9440 \\
Contour traversal reversal & 987 & 809 & \cellcolor{youtuBlue!10}100.00 & \cellcolor{youtuBlue!10}99.88 & 0.2911 \\
Shift + reversal & 987 & 809 & \cellcolor{youtuBlue!10}100.00 & \cellcolor{youtuBlue!10}100.00 & 0.0451 \\
\midrule
\multicolumn{6}{l}{\textbf{(b) Natural-pool comparisons across different construction vectors}}\\[2pt]
Pair population & Pairs & Exact 280D (\%) & Key agreement (\%) & $E_{\mathrm{full}}\!\leq\!1\%$ (\%) & $E_{\mathrm{full}}\!\leq\!5\%$ (\%)\\
\midrule
Distinct \texttt{cad\_vec} & 487,473 & 0.0542 & 0.1061 & 0.1520 & 1.0700 \\
\bottomrule
\end{tabular}
\end{adjustbox}
\caption{\textbf{Complete-descriptor stability and natural-pool collision rates.} (a) Certified counts denote models passing B-rep consistency checks; Compared excludes contour identity cases. (b) Different raw construction vectors form comparison pairs with unassigned geometric equivalence. Exact 280D denotes bytewise equality; Key agreement denotes equality of the quantized keys defined in Section~\ref{app:assembly_conditions}. All rates and maximum errors are percentages.}
\label{tab:full_descriptor_stability}
\end{table}

\begin{figure}[t]
\centering
\includegraphics[width=\textwidth]{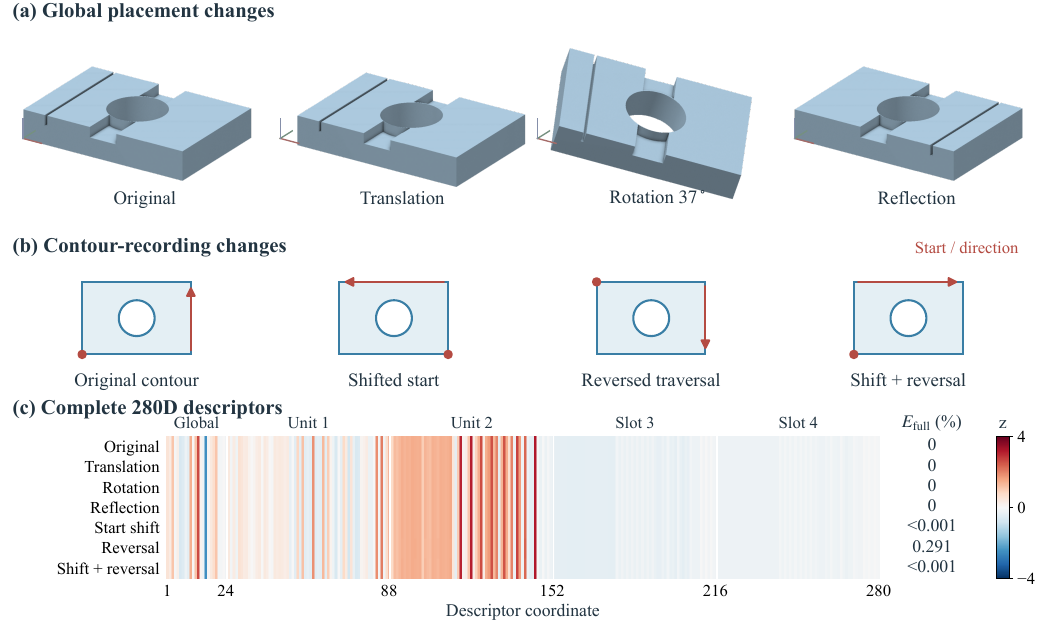}
\caption{\textbf{Complete-descriptor stability for a real Text2CAD test model.} (a) The reconstructed solid under global translation, $37^\circ$ rotation, and reflection in a shared world frame. (b) Its profile under cyclic starting-point shift, traversal reversal, and their combination; red marks show the start and traversal direction. (c) All 280 coordinates use one color scale based on training-set standardization. The coordinate maps and $E_{\mathrm{full}}$ values compare each transformed descriptor with the original. Global placement changes preserve the unit-local inputs; contour variants are recomputed from edited records.}
\label{fig:full_descriptor_example}
\end{figure}

The three global transformations yield identical descriptors. Contour changes achieve a 99.88\%--100\% pass rate at the 1\% threshold and 99.75\%--100\% quantized-key agreement. The few deviations arise from numerical rasterization and area-based slot ordering; rounding can map nearby standardized values to different integers.
Figure~\ref{fig:full_descriptor_example} visualizes this stability across all 280 coordinates for a real test example.

We also compare 487,473 pairs with different raw construction vectors among the reconstructed models. Exact descriptor equality and quantized-key agreement occur in 0.0542\% and 0.1061\% of pairs, respectively. Together, the high agreement under the tested transformations and low cross-record collision rates demonstrate a stable representation that preserves distinctions across the sampled construction records.

\subsection{Descriptor distinctness and pairwise distances}
For the same 4,000-model sample, within-duplicate distances have mean 0.004 and 95th percentile 0.000; between-model distances have mean 20.18 and 5th percentile 6.24. The distance distributions place duplicates close together while separating the sampled models.

\section{Algebraic structure of recording transformations}
\label{app:gauge}
\paragraph{Contours and unit reindexing.}
For a closed contour with $n$ recorded positions, origin shifts form the cyclic group $C_n$; including traversal reversal gives the dihedral group $D_n$. Reindexing $m$ extrusion units corresponds to $S_m$ when unit geometry, Boolean relations, and operation dependencies are preserved.
\paragraph{Construction-record transformations.}
The notation $G=\ZZ_n\times S_k\times B_3\times\mathcal T$ collects contour-start shifts, permutations of $k$ loops or faces, axis conventions, and commuting-operation traces.
The axis factor $B_3=\ZZ_2^3\rtimes S_3$ is the 48-element hyperoctahedral group of signed permutations of three axes. The trace monoid $\mathcal T$ describes commuting Boolean operations, with Foata normal form providing canonical representatives. The notation thus combines group actions with the structure of operation traces.

\section{Training details}
\label{app:training_details}
\subsection{Optimization and model configuration}
The main retrieval model adapts Qwen3-Embedding-0.6B with LoRA using the symmetric contrastive objective in Eq.~\eqref{eq:contrastive_loss}. The Frozen configuration fixes the same text backbone.

Training runs for 30 epochs (18,210 steps) on 8 GPUs, with 128 examples per GPU and a global batch size of 1,024. We use OneCycleLR with learning rates of $10^{-4}$ for the text backbone and $3\times10^{-4}$ for the projection head, together with gradient checkpointing and FP32 precision.

The supplementary 4B LoRA configuration has 77.4M trainable parameters: 70.3M in text LoRA modules and 7.1M in the CAD tower.

\subsection{Baseline implementation and comparison protocol}
\label{app:baseline_settings}
\paragraph{Data and textual inputs.}
All trainable baselines and SpectraBRep are retrained separately on the Text2CAD and MMCAD-B training sets. They share model-level training, validation, and test splits, with all descriptions of a CAD model assigned to one split. Text2CAD v1.0 uses the official splits and all four description levels (Abstract, Beginner, Intermediate, and Expert) for training, with Expert descriptions for the main evaluation. MMCAD-B uses disjoint training, validation, and test models and the Description field as text input.
Each method receives the same original queries, with tokenization, length limits, and truncation determined by its text encoder.

\paragraph{Implementations and initialization.}
CLIP~\citep{radford2021learning} and BLIP-2~\citep{li2023blip} use public pretrained weights for zero-shot evaluation on rendered CAD views, without target-dataset training. The trainable baselines are TriCoLo~\citep{ruan2024tricolo}, RoMa~\citep{feng2025roma}, Parts2Words~\citep{tang2023parts2words}, SCA3D~\citep{ren2025sca3d}, IMMA~\citep{pan2026texttocad}, BRepCLIP~\citep{usama2026brepclip}, Duoduo CLIP~\citep{lee2025duoduoclip}, CLIP-GS~\citep{jiao2025clipgs}, and coordinate-aware CADIR~\citep{liu2026cadir}.

The Bi-GRU text encoders in TriCoLo, Parts2Words, and SCA3D are randomly initialized and fully trained. Text branches based on CLIP, BERT, or Qwen load their respective public pretrained weights and are adapted with LoRA. Existing geometric pretraining is also retained, including the CLIP visual weights in Duoduo CLIP and Uni3D weights in CLIP-GS; the corresponding backbones are updated through LoRA. Other geometric encoders and newly introduced modules are trained from scratch. All LoRA adaptations use identical rank, alpha, and dropout settings.

SpectraBRep keeps its analytically computed 280-dimensional spectral descriptors fixed, trains a randomly initialized residual MLP, and adapts Qwen3-Embedding-0.6B with LoRA.

\paragraph{Geometric inputs and adaptation.}
All inputs come from the same CAD models: rendered views for image methods, surface samples for point methods, voxelized geometry for voxel methods, and faces, edges, and geometric attributes for native B-rep methods. BRepCLIP retains its face--edge encoding, CLIP-GS uses 3D Gaussians, IMMA uses procedural sequences, and CADIR uses construction graphs. All methods use the same test-model list, retaining difficult models in the common evaluation set.

Our implementations preserve TriCoLo's trimodal alignment with its image and voxel branches, Parts2Words' part--word matching, SCA3D's paired data augmentation, Duoduo CLIP's cross-view attention, and IMMA's training-time feature reconstruction.

\paragraph{Training and checkpoint selection.}
Trainable methods use the same model--text training pairs and comparable opportunities for validation-based tuning. Optimizers, learning rates, batch sizes, and training durations follow the original configurations and validation performance. Validation R@5 selects checkpoints; the test set is reserved for evaluation.

We perform target-dataset adaptation separately for Text2CAD and MMCAD-B. Our MMCAD-B results use the checkpoint after three epochs. Direct cross-dataset transfer and training within the target dataset are reported separately.

\paragraph{Evaluation and statistics.}
Following \citet{liu2026cadir}, each query is evaluated against one target model and 999 randomly sampled distractors, giving 1,000 candidates. On both Text2CAD and MMCAD-B, we evaluate with ten randomly selected seeds that control distractor sampling. For each seed, all methods use identical test queries and candidate pools; each method's model weights remain fixed across the ten evaluations.

For both datasets, recall is reported as mean $\pm$ standard deviation across these ten evaluations, reflecting candidate-pool sampling variability. All methods use the same positive-match definitions and tie-handling rules, computing R@1, R@5, R@10, and MedR from per-query rankings. MedR is an integer summary.

Appendix~\ref{app:published_baselines} summarizes additional published results under their original datasets, gallery sizes, and metric definitions. Table~\ref{tab:main} reports our Text2CAD and MMCAD-B experiments.

\Needspace{8\baselineskip}
\section{Additional ablation results}
\label{app:additional_ablations}
\subsection{Descriptor-component visualization}
\begin{figure}[!htbp]
\centering
\includegraphics[width=\linewidth]{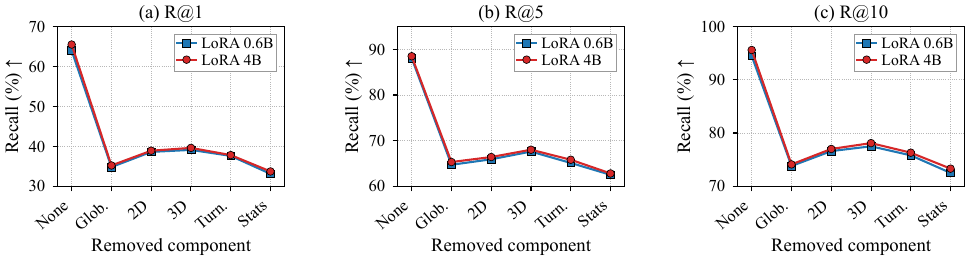}
\caption{\textbf{Retrieval under descriptor-component removal.} Expert recall with LoRA-adapted 0.6B (blue) and 4B (red) text encoders (Table~\ref{tab:geometric_ablation}). Panels show R@1, R@5, and R@10. None denotes Full; other labels identify the removed component. Full has the highest recall at every rank. Local-statistics removal gives the largest R@5 reduction: 25.6 percentage points at 0.6B and 25.7 at 4B.}
\label{fig:geometric_ablation}
\end{figure}

Table~\ref{tab:geometric_ablation} and Section~\ref{sec:geometric_information} report all five descriptor-component ablations. Figure~\ref{fig:geometric_ablation} compares their Expert recall using the same scale within each metric for both LoRA-adapted text backbones.

\subsection{Language-model adaptation across description levels}
The Full descriptor is held fixed while the 0.6B text backbone is either frozen or adapted with LoRA. Table~\ref{tab:small_model_tuning} reports all four description levels under the main evaluation protocol. For Expert descriptions, R@5 increases from 0.618 with Frozen encoding to 0.881 with LoRA. Abstract R@5 is 0.156/0.190 for Frozen/LoRA.

\begin{table}[H]

\vspace{0pt}

\centering
\small
\setlength{\tabcolsep}{2pt}
\renewcommand{\arraystretch}{1.0}
\setlength{\heavyrulewidth}{0.7pt}
\setlength{\lightrulewidth}{0.35pt}
\setlength{\cmidrulewidth}{0.35pt}
\setlength{\aboverulesep}{0.3ex}
\setlength{\belowrulesep}{0.45ex}
\begin{adjustbox}{max width=\linewidth}
\begin{tabular}{@{}llcccc@{}}
\toprule
Level & Tuning & R@1$\uparrow$ & R@5$\uparrow$ & R@10$\uparrow$ & MedR$\downarrow$ \\
\midrule
Abstract & Frozen & \cellcolor{youtuBlue!4.1}0.044 & \cellcolor{youtuBlue!4.1}0.156 & \cellcolor{youtuBlue!4.0}0.233 & \cellcolor{youtuBlue!8.4}44 \\
Abstract & LoRA & \cellcolor{youtuBlue!4.5}\textbf{0.055} & \cellcolor{youtuBlue!5.0}\textbf{0.190} & \cellcolor{youtuBlue!4.9}\textbf{0.266} & \cellcolor{youtuBlue!11.3}\textbf{36} \\
\addlinespace[2pt]
Beginner & Frozen & \cellcolor{youtuBlue!4.0}0.041 & \cellcolor{youtuBlue!4.0}0.153 & \cellcolor{youtuBlue!4.2}0.241 & \cellcolor{youtuBlue!4.0}56 \\
Beginner & LoRA & \cellcolor{youtuBlue!4.8}\textbf{0.065} & \cellcolor{youtuBlue!4.7}\textbf{0.180} & \cellcolor{youtuBlue!5.1}\textbf{0.273} & \cellcolor{youtuBlue!8.7}\textbf{43} \\
\addlinespace[2pt]
Intermediate & Frozen & \cellcolor{youtuBlue!8.7}0.182 & \cellcolor{youtuBlue!11.0}0.406 & \cellcolor{youtuBlue!12.4}0.531 & \cellcolor{youtuBlue!21.1}9 \\
Intermediate & LoRA & \cellcolor{youtuBlue!9.5}\textbf{0.205} & \cellcolor{youtuBlue!13.5}\textbf{0.497} & \cellcolor{youtuBlue!15.0}\textbf{0.623} & \cellcolor{youtuBlue!22.2}\textbf{6} \\
\addlinespace[2pt]
Expert & Frozen & \cellcolor{youtuBlue!14.1}0.344 & \cellcolor{youtuBlue!16.8}0.618 & \cellcolor{youtuBlue!18.0}0.731 & \cellcolor{youtuBlue!23.3}3 \\
Expert & LoRA & \cellcolor{youtuBlue!24.0}\textbf{0.640} & \cellcolor{youtuBlue!24.0}\textbf{0.881} & \cellcolor{youtuBlue!24.0}\textbf{0.945} & \cellcolor{youtuBlue!24.0}\textbf{1} \\
\bottomrule
\end{tabular}
\end{adjustbox}
\caption{\textbf{Frozen versus LoRA text encoding at 0.6B.} Recall scores are mean proportions across ten evaluation seeds for candidate-pool sampling, with 1,000 trials per description level per seed (pool=1,000). Bold marks the better score within each level. Both strategies use the Full descriptor. \emph{Color scale:} darker blue denotes higher recall or lower MedR; each metric shares one scale across both encoders and all description levels.}
\label{tab:small_model_tuning}
\end{table}

\subsection{Number of retained extrusion units}
\label{app:unit_capacity}
The local descriptor in Section~\ref{sec:descriptor} retains a feature vector for each selected extrusion unit. Varying their number $U$ tests whether additional local geometry improves retrieval enough to justify a larger descriptor. Appendix~\ref{app:retained_units} gives the selection and padding rules.

\begin{figure}[!htbp]
\centering
\includegraphics[width=\linewidth]{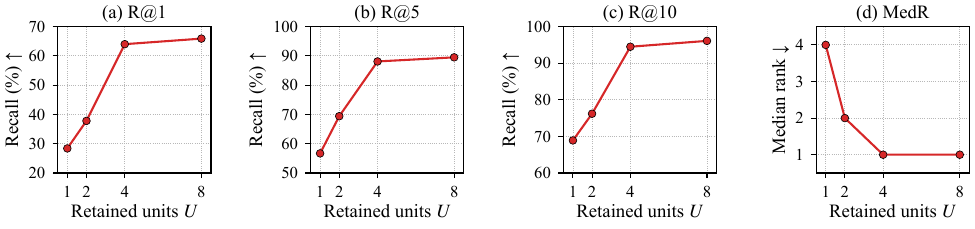}
\caption{\textbf{Number of retained extrusion units.} Text2CAD Expert retrieval with the LoRA-adapted 0.6B text encoder, corresponding to Table~\ref{tab:unit_capacity}. Panels show R@1, R@5, R@10, and MedR. Increasing $U$ from one to four improves all recall metrics and reduces MedR from 4 to 1. Eight units achieve the highest recall with unchanged MedR. The default $U=4$ uses 280 coordinates, 47.8\% fewer than the 536 at $U=8$, balancing retrieval performance and descriptor size (1,000 trials; 1,000 candidates).}
\label{fig:unit_capacity}
\end{figure}

Figure~\ref{fig:unit_capacity} visualizes the LoRA-adapted 0.6B results in Table~\ref{tab:unit_capacity}. Increasing $U$ from one to four improves all three recall metrics and reduces MedR from 4 to 1. Eight units further improve R@1, R@5, and R@10 by 1.9, 1.4, and 1.6 percentage points, respectively, with MedR remaining at 1, while increasing the dimension from 280 to 536.

\subsection{Effect of the 4B text backbone}
\label{app:backbone_reference}
Table~\ref{tab:backbone_reference} reports retrieval results with the 4B text backbone under the same experimental protocol. The corresponding 0.6B Expert results are reported in Table~\ref{tab:main}; Table~\ref{tab:backbone_description_levels} gives all four description levels for 4B LoRA.

\begin{table}[H]

\vspace{0pt}

\centering
\small
\setlength{\tabcolsep}{2pt}
\begin{adjustbox}{max width=\linewidth}
\begin{tabular}{@{}llcccc@{}}
\toprule
Dataset & Tuning & R@1$\uparrow$ & R@5$\uparrow$ & R@10$\uparrow$ & MedR$\downarrow$ \\
\midrule
Text2CAD & LoRA & 0.655 & 0.885 & 0.956 & 1 \\
Text2CAD & Frozen & 0.423 & 0.721 & 0.803 & 2 \\
MMCAD-B & LoRA & 0.399 & 0.688 & 0.776 & 2 \\
\bottomrule
\end{tabular}
\end{adjustbox}
\caption{\textbf{Retrieval with the 4B text backbone.} Qwen3-Embedding-4B results with Text2CAD Expert descriptions and MMCAD-B Description texts. The LoRA system has 4.1B total parameters.}
\label{tab:backbone_reference}
\end{table}

\begin{table}[H]

\vspace{0pt}

\centering
\small
\setlength{\tabcolsep}{7pt}
\begin{adjustbox}{max width=\linewidth}
\begin{tabular}{@{}lccc@{}}
\toprule
Level & R@1$\uparrow$ & R@5$\uparrow$ & R@10$\uparrow$ \\
\midrule
Abstract & 0.063 & 0.181 & 0.266 \\
Beginner & 0.073 & 0.214 & 0.289 \\
Intermediate & 0.261 & 0.523 & 0.636 \\
Expert & \textbf{0.655} & \textbf{0.885} & \textbf{0.956} \\
\bottomrule
\end{tabular}
\end{adjustbox}
\caption{\textbf{Text2CAD description-level recall with 4B LoRA.} Mean recall proportions across ten evaluation seeds are reported using candidate pools of 1,000 models.}
\label{tab:backbone_description_levels}
\end{table}

\section{Published results of additional methods}
\label{app:published_baselines}
Tables~\ref{tab:published_text_retrieval}--\ref{tab:published_brt} collect published results under each paper's original datasets, gallery sizes, and metric definitions. All values are percentages.

BRepCLIP~\citep{usama2026brepclip} uses full galleries of 91K ABC, 40K CADParser, and 65K Automate models. Duoduo CLIP~\citep{lee2025duoduoclip}, trained on Ensembled, uses Text2Shape's approximately 1.4K shapes and 7.4K captions. CLIP-GS~\citep{jiao2025clipgs} uses 20K Objaverse-GS shapes, with Objaverse-LVIS excluded from training.

\begin{table}[H]

\vspace{0pt}

\centering
\small
\setlength{\tabcolsep}{3pt}

\begin{adjustbox}{max width=\linewidth}
\begin{tabular}{@{}llrrr@{}}
\toprule
Method & Test dataset & @1 & @5 & @10 \\
\midrule
BRepCLIP & ABC & 8.59 & 24.52 & 35.08 \\
 & CADParser & 5.00 & 15.08 & 22.12 \\
 & Automate & 9.42 & 24.18 & 32.86 \\
Duoduo CLIP & Text2Shape & 15.90 & 34.08 & -- \\
CLIP-GS & Objaverse-GS & 36.8 & 68.1 & 79.9 \\
\bottomrule
\end{tabular}
\end{adjustbox}
\caption{Published text-to-shape retrieval: BRepCLIP Top-$k$ accuracy (source Table~1), Duoduo CLIP RR@$k$ (Table~4), and CLIP-GS R@$k$ (Table~1). Values are percentages; a dash denotes an unreported score.}
\label{tab:published_text_retrieval}
\end{table}

\begin{samepage}
CADGCL~\citep{qin2025cadgcl} reports CAD-to-CAD retrieval with an 80:20 train/test split. BRT~\citep{zou2025brt} reports classification and face labeling with 70:15:15 train/validation/test splits. Applying either encoder to text-to-CAD retrieval requires an additional text-alignment stage.\par
\end{samepage}

\begin{table}[H]

\vspace{0pt}

\centering
\small
\setlength{\tabcolsep}{5pt}

\begin{adjustbox}{max width=\linewidth}
\begin{tabular}{@{}lrr@{}}
\toprule
Test dataset & mAP (top 10) & mAP (top 50) \\
\midrule
BRepCAD30 & 94.67 & 84.51 \\
FabWave & 98.58 & 89.35 \\
\bottomrule
\end{tabular}
\end{adjustbox}
\caption{Published CADGCL CAD-to-CAD retrieval (source Tables~1--2).}
\label{tab:published_cadgcl}
\end{table}

\begin{table}[H]

\vspace{0pt}

\centering
\small
\setlength{\tabcolsep}{2pt}

\begin{adjustbox}{max width=\linewidth}
\begin{tabular}{@{}llrr@{}}
\toprule
Task & Dataset & Accuracy & Mean IoU \\
\midrule
\multirow{3}{*}{Part classif.} & TMCAD & 83.45 & -- \\
 & FabWave & 98.95 & -- \\
 & SolidLetters & 97.35 & -- \\
\midrule
Feature recog. & MFCAD++ & 99.27 & 97.98 \\
\midrule
Op. segment. & Fusion 360 Gallery & 94.48 & 79.23 \\
\bottomrule
\end{tabular}
\end{adjustbox}
\caption{Published BRT results (source Tables~1--3): part classification, machining-feature recognition, and modeling-operation segmentation. A dash denotes inapplicable IoU.}
\label{tab:published_brt}
\end{table}

%% file: references.bib
@inproceedings{khan2024text2cad,
  title = {{Text2CAD: Generating Sequential CAD Designs from Beginner-to-Expert Level Text Prompts}},
  author = {Khan, Mohammad Sadil and Sinha, Sankalp and Sheikh, Talha Uddin and Stricker, Didier and Ali, Sk Aziz and Afzal, Muhammad Zeshan},
  year = {2024},
  booktitle = {Advances in Neural Information Processing Systems},
  volume = {37},
  pages = {7552--7579},
}

@article{kac1966can,
  title = {{Can One Hear the Shape of a Drum?}},
  author = {Kac, Mark},
  year = {1966},
  journal = {The American Mathematical Monthly},
  volume = {73},
  number = {4P2},
  pages = {1--23},
}

@article{arkin1991efficiently,
  title = {{An Efficiently Computable Metric for Comparing Polygonal Shapes}},
  author = {Arkin, Esther M. and Chew, L. Paul and Huttenlocher, Daniel P. and Kedem, Klara and Mitchell, Joseph S. B.},
  year = {1991},
  journal = {IEEE Transactions on Pattern Analysis and Machine Intelligence},
  volume = {13},
  number = {3},
  pages = {209--216},
}

@inproceedings{wu2021deepcad,
  title = {{DeepCAD: A Deep Generative Network for Computer-Aided Design Models}},
  author = {Wu, Rundi and Xiao, Chang and Zheng, Changxi},
  year = {2021},
  booktitle = {Proceedings of the IEEE/CVF International Conference on Computer Vision},
  pages = {6772--6782},
}

@inproceedings{xu2022skexgen,
  title = {{SkexGen: Autoregressive Generation of CAD Construction Sequences with Disentangled Codebooks}},
  author = {Xu, Xiang and Willis, Karl D. D. and Lambourne, Joseph G. and Cheng, Chin-Yi and Jayaraman, Pradeep Kumar and Furukawa, Yasutaka},
  year = {2022},
  booktitle = {International Conference on Machine Learning},
  volume = {162},
  series = {Proceedings of Machine Learning Research},
  pages = {24698--24724},
}

@article{khan2024cadsignet,
  title = {{CAD-SIGNet: CAD Language Inference from Point Clouds using Layer-wise Sketch Instance Guided Attention}},
  author = {Khan, Mohammad Sadil and Dupont, Elona and Ali, Sk Aziz and Cherenkova, Kseniya and Kacem, Anis and Aouada, Djamila},
  year = {2024},
  journal = {arXiv preprint arXiv:2402.17678v1},
  url = {https://arxiv.org/abs/2402.17678v1}
}

@inproceedings{achlioptas2019shapeglot,
  title = {{Shapeglot: Learning Language for Shape Differentiation}},
  author = {Achlioptas, Panos and Fan, Judy and Hawkins, Robert and Goodman, Noah and Guibas, Leonidas J.},
  year = {2019},
  booktitle = {Proceedings of the IEEE/CVF International Conference on Computer Vision},
  pages = {8938--8947},
}

@inproceedings{chen2023text2shape,
  title = {{Text2Shape: Generating Shapes from Natural Language by Learning Joint Embeddings}},
  author = {Chen, Kevin and Choy, Christopher B. and Savva, Manolis and Chang, Angel X. and Funkhouser, Thomas and Savarese, Silvio},
  year = {2019},
  booktitle = {Computer Vision -- ACCV 2018},
  volume = {11363},
  series = {Lecture Notes in Computer Science},
  pages = {100--116},
}

@article{xue2023ulip,
  title = {{ULIP: Learning a Unified Representation of Language, Images, and Point Clouds for 3D Understanding}},
  author = {Xue, Le and Gao, Mingfei and Xing, Chen and Mart{\'\i}n-Mart{\'\i}n, Roberto and Wu, Jiajun and Xiong, Caiming and Xu, Ran and Niebles, Juan Carlos and Savarese, Silvio},
  year = {2023},
  journal = {arXiv preprint arXiv:2212.05171v4},
  url = {https://arxiv.org/abs/2212.05171v4}
}

@inproceedings{liu2024openshape,
  title = {{OpenShape: Scaling Up 3D Shape Representation Towards Open-World Understanding}},
  author = {Liu, Minghua and Shi, Ruoxi and Kuang, Kaiming and Zhu, Yinhao and Li, Xuanlin and Han, Shizhong and Cai, Hong and Porikli, Fatih and Su, Hao},
  year = {2023},
  booktitle = {Advances in Neural Information Processing Systems},
  volume = {36},
}

@article{reuter2006laplace,
  title = {{Laplace--Beltrami Spectra as `Shape-DNA' of Surfaces and Solids}},
  author = {Reuter, Martin and Wolter, Franz-Erich and Peinecke, Niklas},
  year = {2006},
  journal = {Computer-Aided Design},
  volume = {38},
  number = {4},
  pages = {342--366},
}

@article{sun2009concise,
  title = {{A Concise and Provably Informative Multi-Scale Signature Based on Heat Diffusion}},
  author = {Sun, Jian and Ovsjanikov, Maks and Guibas, Leonidas},
  year = {2009},
  journal = {Computer Graphics Forum},
  volume = {28},
  number = {5},
  pages = {1383--1392},
}

@inproceedings{aubry2011wave,
  title = {{The Wave Kernel Signature: A Quantum Mechanical Approach to Shape Analysis}},
  author = {Aubry, Mathieu and Schlickewei, Ulrich and Cremers, Daniel},
  year = {2011},
  booktitle = {IEEE International Conference on Computer Vision Workshops},
  pages = {1626--1633},
}

@inproceedings{radford2021learning,
  title = {{Learning Transferable Visual Models From Natural Language Supervision}},
  author = {Radford, Alec and Kim, Jong Wook and Hallacy, Chris and Ramesh, Aditya and Goh, Gabriel and Agarwal, Sandhini and Sastry, Girish and Askell, Amanda and Mishkin, Pamela and Clark, Jack and Krueger, Gretchen and Sutskever, Ilya},
  year = {2021},
  booktitle = {International Conference on Machine Learning},
  volume = {139},
  series = {Proceedings of Machine Learning Research},
  pages = {8748--8763},
}

@inproceedings{li2023blip,
  title = {{BLIP-2: Bootstrapping Language-Image Pre-training with Frozen Image Encoders and Large Language Models}},
  author = {Li, Junnan and Li, Dongxu and Savarese, Silvio and Hoi, Steven},
  year = {2023},
  booktitle = {International Conference on Machine Learning},
  volume = {202},
  series = {Proceedings of Machine Learning Research},
  pages = {19730--19742},
}

@article{requicha1980solids,
  title = {{Representations for Rigid Solids: Theory, Methods, and Systems}},
  author = {Requicha, Aristides A. G.},
  year = {1980},
  journal = {ACM Computing Surveys},
  volume = {12},
  number = {4},
  pages = {437--464},
}

@article{lambourne2021brepnet,
  title = {{BRepNet: A topological message passing system for solid models}},
  author = {Lambourne, Joseph G. and Willis, Karl D. D. and Jayaraman, Pradeep Kumar and Sanghi, Aditya and Meltzer, Peter and Shayani, Hooman},
  year = {2021},
  journal = {arXiv preprint arXiv:2104.00706v2},
  url = {https://arxiv.org/abs/2104.00706v2}
}

@article{jayaraman2021uvnet,
  title = {{UV-Net: Learning from Boundary Representations}},
  author = {Jayaraman, Pradeep Kumar and Sanghi, Aditya and Lambourne, Joseph G. and Willis, Karl D. D. and Davies, Thomas and Shayani, Hooman and Morris, Nigel},
  year = {2021},
  journal = {arXiv preprint arXiv:2006.10211v2},
  url = {https://arxiv.org/abs/2006.10211v2}
}

@article{qi2017pointnet,
  title = {{PointNet: Deep Learning on Point Sets for 3D Classification and Segmentation}},
  author = {Qi, Charles R. and Su, Hao and Mo, Kaichun and Guibas, Leonidas J.},
  year = {2017},
  journal = {arXiv preprint arXiv:1612.00593v2},
  url = {https://arxiv.org/abs/1612.00593v2}
}

@inproceedings{qi2017pointnetpp,
  title = {{PointNet++: Deep Hierarchical Feature Learning on Point Sets in a Metric Space}},
  author = {Qi, Charles R. and Yi, Li and Su, Hao and Guibas, Leonidas J.},
  year = {2017},
  booktitle = {Advances in Neural Information Processing Systems},
  volume = {30},
}

@article{wang2019dgcnn,
  title = {{Dynamic Graph CNN for Learning on Point Clouds}},
  author = {Wang, Yue and Sun, Yongbin and Liu, Ziwei and Sarma, Sanjay E. and Bronstein, Michael M. and Solomon, Justin M.},
  year = {2019},
  journal = {ACM Transactions on Graphics},
  volume = {38},
  number = {5},
  pages = {146:1--146:12},
}

@article{hanocka2019meshcnn,
  title = {{MeshCNN: A Network with an Edge}},
  author = {Hanocka, Rana and Hertz, Amir and Fish, Noa and Giryes, Raja and Fleishman, Shachar and Cohen-Or, Daniel},
  year = {2019},
  journal = {ACM Transactions on Graphics},
  volume = {38},
  number = {4},
  pages = {90:1--90:12},
}

@article{wu2015shapenets,
  title = {{3D ShapeNets: A Deep Representation for Volumetric Shapes}},
  author = {Wu, Zhirong and Song, Shuran and Khosla, Aditya and Yu, Fisher and Zhang, Linguang and Tang, Xiaoou and Xiao, Jianxiong},
  year = {2015},
  journal = {arXiv preprint arXiv:1406.5670v3},
  url = {https://arxiv.org/abs/1406.5670v3}
}

@inproceedings{su2015multiview,
  title = {{Multi-View Convolutional Neural Networks for 3D Shape Recognition}},
  author = {Su, Hang and Maji, Subhransu and Kalogerakis, Evangelos and Learned-Miller, Erik},
  year = {2015},
  booktitle = {Proceedings of the IEEE International Conference on Computer Vision},
  pages = {945--953},
}

@article{chang2015shapenet,
  title = {{ShapeNet: An Information-Rich 3D Model Repository}},
  author = {Chang, Angel X. and Funkhouser, Thomas and Guibas, Leonidas and Hanrahan, Pat and Huang, Qixing and Li, Zimo and Savarese, Silvio and Savva, Manolis and Song, Shuran and Su, Hao and Xiao, Jianxiong and Yi, Li and Yu, Fisher},
  year = {2015},
  journal = {arXiv preprint arXiv:1512.03012},
}

@article{koch2019abc,
  title = {{ABC: A Big CAD Model Dataset For Geometric Deep Learning}},
  author = {Koch, Sebastian and Matveev, Albert and Jiang, Zhongshi and Williams, Francis and Artemov, Alexey and Burnaev, Evgeny and Alexa, Marc and Zorin, Denis and Panozzo, Daniele},
  year = {2019},
  journal = {arXiv preprint arXiv:1812.06216v2},
  url = {https://arxiv.org/abs/1812.06216v2}
}

@article{willis2021fusion,
  title = {{Fusion 360 Gallery: A Dataset and Environment for Programmatic CAD Construction from Human Design Sequences}},
  author = {Willis, Karl D. D. and Pu, Yewen and Luo, Jieliang and Chu, Hang and Du, Tao and Lambourne, Joseph G. and Solar-Lezama, Armando and Matusik, Wojciech},
  year = {2021},
  journal = {ACM Transactions on Graphics},
  volume = {40},
  number = {4},
  pages = {54:1--54:24},
}

@inproceedings{zhou2024uni3d,
  title = {{Uni3D: Exploring Unified 3D Representation at Scale}},
  author = {Zhou, Junsheng and Wang, Jinsheng and Ma, Baorui and Liu, Yu-Shen and Huang, Tiejun and Wang, Xinlong},
  year = {2024},
  booktitle = {International Conference on Learning Representations},
}

@article{xue2024ulip2,
  title = {{ULIP-2: Towards Scalable Multimodal Pre-training for 3D Understanding}},
  author = {Xue, Le and Yu, Ning and Zhang, Shu and Panagopoulou, Artemis and Li, Junnan and Mart{\'\i}n-Mart{\'\i}n, Roberto and Wu, Jiajun and Xiong, Caiming and Xu, Ran and Niebles, Juan Carlos and Savarese, Silvio},
  year = {2024},
  journal = {arXiv preprint arXiv:2305.08275v4},
  url = {https://arxiv.org/abs/2305.08275v4}
}

@article{zhang2022pointclip,
  title = {{PointCLIP: Point Cloud Understanding by CLIP}},
  author = {Zhang, Renrui and Guo, Ziyu and Zhang, Wei and Li, Kunchang and Miao, Xupeng and Cui, Bin and Qiao, Yu and Gao, Peng and Li, Hongsheng},
  year = {2021},
  journal = {arXiv preprint arXiv:2112.02413v1},
  url = {https://arxiv.org/abs/2112.02413v1}
}

@article{liu2026cadir,
  title = {{CADIR: A Cross-Backend Editable Intermediate Representation for Agentic CAD Generation}},
  author = {Liu, Yu and Ni, Jingzhe and Chen, Yiming and Huang, Junqi and Tong, Ruofeng and Tang, Min and Du, Peng},
  year = {2026},
  journal = {arXiv preprint arXiv:2608.00891},
}

@article{zahn1972fourier,
  title = {{Fourier Descriptors for Plane Closed Curves}},
  author = {Zahn, Charles T. and Roskies, Ralph Z.},
  year = {1972},
  journal = {IEEE Transactions on Computers},
  volume = {C-21},
  number = {3},
  pages = {269--281},
}

@article{persoon1977fourier,
  title = {{Shape Discrimination Using Fourier Descriptors}},
  author = {Persoon, Eric and Fu, King-Sun},
  year = {1977},
  journal = {IEEE Transactions on Systems, Man, and Cybernetics},
  volume = {SMC-7},
  number = {3},
  pages = {170--179},
}

@article{kuhl1982elliptic,
  title = {{Elliptic Fourier Features of a Closed Contour}},
  author = {Kuhl, Frank P. and Giardina, Charles R.},
  year = {1982},
  journal = {Computer Graphics and Image Processing},
  volume = {18},
  number = {3},
  pages = {236--258},
}

@inproceedings{chen2020simclr,
  title = {{A Simple Framework for Contrastive Learning of Visual Representations}},
  author = {Chen, Ting and Kornblith, Simon and Norouzi, Mohammad and Hinton, Geoffrey},
  year = {2020},
  booktitle = {International Conference on Machine Learning},
  volume = {119},
  series = {Proceedings of Machine Learning Research},
  pages = {1597--1607},
}

@inproceedings{wang2020alignment,
  title = {{Understanding Contrastive Representation Learning through Alignment and Uniformity on the Hypersphere}},
  author = {Wang, Tongzhou and Isola, Phillip},
  year = {2020},
  booktitle = {International Conference on Machine Learning},
  volume = {119},
  series = {Proceedings of Machine Learning Research},
  pages = {9929--9939},
}

@article{jacobson2013winding,
  title = {{Robust Inside-Outside Segmentation using Generalized Winding Numbers}},
  author = {Jacobson, Alec and Kavan, Ladislav and Sorkine-Hornung, Olga},
  year = {2013},
  journal = {ACM Transactions on Graphics},
  volume = {32},
  number = {4},
}

@article{seff2020sketchgraphs,
  title = {{SketchGraphs: A Large-Scale Dataset for Modeling Relational Geometry in Computer-Aided Design}},
  author = {Seff, Ari and Ovadia, Yaniv and Zhou, Wenda and Adams, Ryan P.},
  year = {2020},
  journal = {arXiv preprint arXiv:2007.08506},
}

@inproceedings{zhu2023pointclipv2,
  title = {{PointCLIP V2: Prompting CLIP and GPT for Powerful 3D Open-world Learning}},
  author = {Zhu, Xiangyang and Zhang, Renrui and He, Bowei and Guo, Ziyu and Zeng, Ziyao and Qin, Zipeng and Zhang, Shanghang and Gao, Peng},
  year = {2023},
  booktitle = {Proceedings of the IEEE/CVF International Conference on Computer Vision},
  pages = {2639--2650},
}

@inproceedings{cohen2016equivariant,
  title = {{Group Equivariant Convolutional Networks}},
  author = {Cohen, Taco and Welling, Max},
  year = {2016},
  booktitle = {International Conference on Machine Learning},
  volume = {48},
  series = {Proceedings of Machine Learning Research},
  pages = {2990--2999},
}

@article{peinecke2007fingerprints,
  title = {{Laplace Spectra as Fingerprints for Image Recognition}},
  author = {Peinecke, Niklas and Wolter, Franz-Erich and Reuter, Martin},
  year = {2007},
  journal = {Computer-Aided Design},
  volume = {39},
  number = {6},
  pages = {460--476},
}

@article{gordon1992drum,
  title = {{One Cannot Hear the Shape of a Drum}},
  author = {Gordon, Carolyn and Webb, David L. and Wolpert, Scott},
  year = {1992},
  journal = {Bulletin of the American Mathematical Society},
  volume = {27},
  number = {1},
  pages = {134--138},
}

@article{usama2026brepclip,
  title = {{BRepCLIP}: Contrastive Multimodal Pretraining on {BRep} Primitives for {CAD} Understanding},
  author = {Usama, Muhammad and Stricker, Didier and Khan, Mohammad Sadil and Afzal, Muhammad Zeshan},
  journal = {arXiv preprint arXiv:2606.05515},
  year = {2026},
  url = {https://arxiv.org/abs/2606.05515}
}

@inproceedings{lee2025duoduoclip,
  title = {Duoduo {CLIP}: Efficient {3D} Understanding with Multi-View Images},
  author = {Lee, Han-Hung and Zhang, Yiming and Chang, Angel X.},
  booktitle = {International Conference on Learning Representations},
  year = {2025},
  url = {https://openreview.net/forum?id=iGbuc9ekKK}
}

@inproceedings{jiao2025clipgs,
  title = {{CLIP-GS}: Unifying Vision-Language Representation with {3D} Gaussian Splatting},
  author = {Jiao, Siyu and Dong, Haoye and Yin, Yuyang and Jie, Zequn and Qian, Yinlong and Zhao, Yao and Shi, Humphrey and Wei, Yunchao},
  booktitle = {Proceedings of the IEEE/CVF International Conference on Computer Vision},
  pages = {4670--4680},
  year = {2025}
}

@article{zou2025brt,
  title = {Bringing Attention to {CAD}: Boundary Representation Learning via Transformer},
  author = {Zou, Qiang and Zhu, Lizhen},
  journal = {Computer-Aided Design},
  volume = {189},
  pages = {103940},
  year = {2025},
  doi = {10.1016/j.cad.2025.103940}
}

@article{qin2025cadgcl,
  title = {{CADGCL}: Unsupervised Retrieval of {CAD} Models via Boundary Representations},
  author = {Qin, Feiwei and Zhu, Liangzhe and Xu, Zijian and Fang, Meie and Li, Ping},
  journal = {The Visual Computer},
  volume = {41},
  pages = {6493--6505},
  year = {2025},
  doi = {10.1007/s00371-025-03949-y}
}

@inproceedings{ruan2024tricolo,
  title = {{TriCoLo}: Trimodal Contrastive Loss for Text To Shape Retrieval},
  author = {Ruan, Yue and Lee, Han-Hung and Zhang, Yiming and Zhang, Ke and Chang, Angel X.},
  booktitle = {Proceedings of the IEEE/CVF Winter Conference on Applications of Computer Vision},
  pages = {5815--5825},
  year = {2024},
  url = {https://openaccess.thecvf.com/content/WACV2024/html/Ruan_TriCoLo_Trimodal_Contrastive_Loss_for_Text_To_Shape_Retrieval_WACV_2024_paper.html}
}

@article{feng2025roma,
  title = {{PointCloud-Text} Matching: Benchmark Datasets and a Baseline},
  author = {Feng, Yanglin and Qin, Yang and Peng, Dezhong and Zhu, Hongyuan and Peng, Xi and Hu, Peng},
  journal = {arXiv preprint arXiv:2403.19386},
  year = {2025},
  url = {https://arxiv.org/abs/2403.19386v3}
}

@article{tang2023parts2words,
  title = {{Parts2Words: Learning Joint Embedding of Point Clouds and Texts by Bidirectional Matching between Parts and Words}},
  author = {Tang, Chuan and Yang, Xi and Wu, Bojian and Han, Zhizhong and Chang, Yi},
  year = {2023},
  journal = {arXiv preprint arXiv:2107.01872v2},
  url = {https://arxiv.org/abs/2107.01872v2}
}

@article{ren2025sca3d,
  title = {{SCA3D}: Enhancing Cross-modal {3D} Retrieval via {3D} Shape and Caption Paired Data Augmentation},
  author = {Ren, Junlong and Wu, Hao and Xiong, Hui and Wang, Hao},
  journal = {arXiv preprint arXiv:2502.19128},
  year = {2025},
  url = {https://arxiv.org/abs/2502.19128}
}

@article{pan2026texttocad,
  title = {Text-to-{CAD} Retrieval: a Strong Baseline},
  author = {Pan, Honghu and Du, Zibo and Liu, Daxiang and Liu, Chengliang and Luo, Xiaoling},
  journal = {arXiv preprint arXiv:2605.05572},
  year = {2026},
  url = {https://arxiv.org/abs/2605.05572}
}

@article{jin2026himmel,
  title = {{HIMMEL: Hierarchical Interleaved Multi-stream Motion Encoding for Long Video Understanding}},
  author = {Haopeng Jin and Wenlong Zhao and Tao Yu and Tiankun Yang and Zhenyu Guan and ShiQuan Dong and Yujia Yang and Shani Ye and Chubin Chen and Hongzhu Yi and Jinwen Luo},
  journal = {arXiv preprint arXiv:2605.08158},
  year = {2026},
  url = {https://arxiv.org/abs/2605.08158},
}

@article{zhang2026deltajepa,
  title = {{Delta-JEPA: Learning Action-Sensitive World Models via Latent Difference Decoding}},
  author = {Zhenghao Zhang and Yuanxiang Wang and Zhenyu Guan and Yujia Yang and Bingkang Shi and Tianyu Zong and Hongzhu Yi and Guoqing Chao and Xingchen Chen and Tiankun Yang and Chenxi Bao and Tao Yu and Jingjing Zhou and Jungang Xu},
  journal = {arXiv preprint arXiv:2606.31232},
  year = {2026},
  url = {https://arxiv.org/abs/2606.31232},
}

@article{yu2025browseragent,
  title = {{BrowserAgent: Building Web Agents with Human-Inspired Web Browsing Actions}},
  author = {Tao Yu and Zhengbo Zhang and Zhiheng Lyu and Junhao Gong and Hongzhu Yi and Xinming Wang and Yuxuan Zhou and Jiabing Yang and Ping Nie and Yan Huang and Wenhu Chen},
  journal = {arXiv preprint arXiv:2510.10666},
  year = {2025},
  url = {https://arxiv.org/abs/2510.10666},
}

@article{yu2026entcollabbench,
  title = {{Beyond the All-in-One Agent: Benchmarking Role-Specialized Multi-Agent Collaboration in Enterprise Workflows}},
  author = {Tao Yu and Hao Wang and Changyu Li and Shenghua Chai and Minghui Zhang and Zhongtian Luo and Yuxuan Zhou and Haopeng Jin and Zhaolu Kang and Jiabing Yang and YiFan Zhang and Xinming Wang and Hongzhu Yi and Zheqi He and Jing-Shu Zheng and Xi Yang and Yan Huang and Liang Wang},
  journal = {arXiv preprint arXiv:2605.08761},
  year = {2026},
  url = {https://arxiv.org/abs/2605.08761},
}

@article{yu2026paperx,
  title = {{PaperX: A Unified Framework for Multimodal Academic Presentation Generation with Scholar DAG}},
  author = {Tao Yu and Minghui Zhang and Zhiqing Cui and Hao Wang and Zhongtian Luo and Shenghua Chai and Junhao Gong and Yuzhao Peng and Yuxuan Zhou and Yujia Yang and Zhenghao Zhang and Haopeng Jin and Xinming Wang and Yufei Xiong and Jiabing Yang and Jiahao Yuan and Hanqing Wang and Hongzhu Yi and Yan Huang and Liang Wang},
  journal = {arXiv preprint arXiv:2602.03866},
  year = {2026},
  url = {https://arxiv.org/abs/2602.03866},
}

@article{yi2026manifold,
  title = {{Bridging the Manifold Gap: Riemannian Residual Line Search for One-Step Image Editing}},
  author = {Hongzhu Yi and Zhongtian Luo and Tong Li and Yiyan Fan and Jungang Xu},
  journal = {arXiv preprint arXiv:2606.24844},
  year = {2026},
  url = {https://arxiv.org/abs/2606.24844},
}

@article{yu2026omnideepsearch,
  title = {{Omni-DeepSearch: A Benchmark for Audio-Driven Omni-Modal Deep Search}},
  author = {Tao Yu and Yiming Ding and Shenghua Chai and Minghui Zhang and Zhongtian Luo and Xinming Wang and Xinlong Chen and Zhaolu Kang and Junhao Gong and Yuxuan Zhou and Haopeng Jin and Zhiqing Cui and Jiabing Yang and YiFan Zhang and Hongzhu Yi and Zheqi He and Xi Yang and Yan Huang and Liang Wang},
  journal = {arXiv preprint arXiv:2605.08762},
  year = {2026},
  url = {https://arxiv.org/abs/2605.08762},
}

@article{yi2026rpo,
  title = {{RPO: Reinforcement Fine-Tuning with Partial Reasoning Optimization}},
  author = {Hongzhu Yi and Xinming Wang and Zhenghao Zhang and Tianyu Zong and Yuanxiang Wang and Jun Xie and Tao Yu and Haopeng Jin and Kaixin Xu and Feng Chen and Jiahuan Chen and Yujia Yang and Zhenyu Guan and Bingkang Shi and Jungang Xu},
  journal = {arXiv preprint arXiv:2601.19404},
  year = {2026},
  url = {https://arxiv.org/abs/2601.19404},
}

@article{yi2026vde,
  title = {{VDE Bench: Evaluating The Capability of Image Editing Models to Modify Visual Documents}},
  author = {Hongzhu Yi and Yujia Yang and Yuanxiang Wang and Tong Li and Zhenyu Guan and Tianyu Zong and Jiahuan Chen and Chenxi Bao and Tiankun Yang and Haopeng Jin and Yixuan Yuan and Xinming Wang and Tao Yu and Ruilin Gao and Ruiwen Tao and Haijin Liang and Jin Ma and Jinwen Luo and Yeshani and Xinyu Zuo and Jungang Xu},
  journal = {arXiv preprint arXiv:2602.00122},
  year = {2026},
  url = {https://arxiv.org/abs/2602.00122},
}

@article{yu2026rsebench,
  title = {{When Non-Adversarial Noise Hurts Learning: Benchmarking Skill Self-Evolution in Agents}},
  author = {Tao Yu and Zhongtian Luo and Yuchen Long and Yiheng Lu and Yuxuan Zhou and Yanpei Gong and Siyi Cao and Yuqing Wu and Haopeng Jin and Zhenghao Zhang and Yujia Yang and Jingmin Zhu and Yunyun Hou and Xinming Wang and Hongzhu Yi and Jiachen Liu and Sirui Wang and Nan Duan and Yan Huang and Liang Wang and Tianjin Huang and Yanyan Zou},
  journal = {arXiv preprint arXiv:2610.19910},
  year = {2026},
  url = {https://arxiv.org/abs/2610.19910},
}

@article{yu2026scoperouter,
  title = {{SCOPE-Router: Cost-Aware Open-Set VLM Routing for Execution-Oriented Tasks}},
  author = {Tao Yu and Yifei Qu and Zhiqing Cui and Pengfei Zhou and Zhongtian Luo and Yujia Yang and Shenghua Chai and Haopeng Jin and Zhenghao Zhang and Xinming Wang and Hongzhu Yi and Wangbo Zhao and Zhenglin Wan and Yan Huang and Yeshani and Jinwen Luo and Yang You},
  journal = {arXiv preprint arXiv:2608.12127},
  year = {2026},
  url = {https://arxiv.org/abs/2608.12127},
}

@article{yu2026rvms,
  title = {{Beyond Closed-Pool Video Retrieval: A Benchmark and Agent Framework for Real-World Video Search and Moment Localization}},
  author = {Tao Yu and Yujia Yang and Haopeng Jin and Junhao Gong and Xinlong Chen and Yuxuan Zhou and Shanbin Zhang and Jiabing Yang and Xinming Wang and Hongzhu Yi and Ping Nie and Kai Zou and Zhang Zhang and Yan Huang and Liang Wang and Yeshani and Ruiwen Tao and Jin Ma and Haijin Liang and Jinwen Luo},
  journal = {arXiv preprint arXiv:2602.10159},
  year = {2026},
  url = {https://arxiv.org/abs/2602.10159},
}

@article{wang2026opis,
  title = {{OPIS: An Input-Grounded Benchmark for Multi-Object Memory in Video World Models}},
  author = {Hao Wang and Tao Yu and Liuzhou Zhang and HeXin Wang and Haopeng Jin and Yuxuan Zhou and Xinming Wang and Hongzhu Yi and Xinye Li and Yuanlei Wang and Ping Nie and Yan Huang and Yuxuan Zhang and Pengfei Zhou and Yanyan Zou and Wei Yang},
  journal = {arXiv preprint arXiv:2609.35052},
  year = {2026},
  url = {https://arxiv.org/abs/2609.35052},
}

@article{wu2026spectraldistillation,
  title = {{Towards Principled Dataset Distillation: A Spectral Distribution Perspective}},
  author = {Ruixi Wu and Shaobo Wang and Jiahuan Chen and Zhiyuan Liu and Yicun Yang and Zhaorun Chen and Zekai Li and Kaixin Li and Xinming Wang and Hongzhu Yi and Kai Wang and Linfeng Zhang},
  journal = {arXiv preprint arXiv:2603.01698},
  year = {2026},
  url = {https://arxiv.org/abs/2603.01698},
}

@article{xie2025foureyes,
  title = {{Four Eyes Are Better Than Two: Harnessing the Collaborative Potential of Large Models via Differentiated Thinking and Complementary Ensembles}},
  author = {Jun Xie and Xiongjun Guan and Yingjian Zhu and Zhaoran Zhao and Xinming Wang and Hongzhu Yi and Feng Chen and Zhepeng Wang},
  journal = {arXiv preprint arXiv:2505.16784},
  year = {2025},
  url = {https://arxiv.org/abs/2505.16784},
}

@article{xie2025more,
  title = {{More Is Better: A MoE-Based Emotion Recognition Framework with Human Preference Alignment}},
  author = {Jun Xie and Yingjian Zhu and Feng Chen and Zhenghao Zhang and Xiaohui Fan and Hongzhu Yi and Xinming Wang and Chen Yu and Yue Bi and Zhaoran Zhao and Xiongjun Guan and Zhepeng Wang},
  journal = {arXiv preprint arXiv:2508.06036},
  year = {2025},
  url = {https://arxiv.org/abs/2508.06036},
}

@article{yang2026omniiie,
  title = {{Omni IIE Bench: Benchmarking the Practical Capabilities of Image Editing Models}},
  author = {Yujia Yang and Yuanxiang Wang and Zhenyu Guan and Tiankun Yang and Chenxi Bao and Haopeng Jin and Jinwen Luo and Xinyu Zuo and Lisheng Duan and Haijin Liang and Jin Ma and Xinming Wang and Ruiwen Tao and Hongzhu Yi},
  journal = {arXiv preprint arXiv:2603.16944},
  year = {2026},
  url = {https://arxiv.org/abs/2603.16944},
}

@article{zong2025jtcse,
  title = {{JTCSE: Joint Tensor-Modulus Constraints and Cross-Attention for Unsupervised Contrastive Learning of Sentence Embeddings}},
  author = {Tianyu Zong and Hongzhu Yi and Bingkang Shi and Yuanxiang Wang and Jungang Xu},
  journal = {arXiv preprint arXiv:2505.02366},
  year = {2025},
  url = {https://arxiv.org/abs/2505.02366},
}

@article{zhang2025dgimvcm,
  title = {{Dynamic Deep Graph Learning for Incomplete Multi-View Clustering with Masked Graph Reconstruction Loss}},
  author = {Zhenghao Zhang and Jun Xie and Xingchen Chen and Tao Yu and Hongzhu Yi and Kaixin Xu and Yuanxiang Wang and Tianyu Zong and Xinming Wang and Jiahuan Chen and Guoqing Chao and Feng Chen and Zhepeng Wang and Jungang Xu},
  journal = {arXiv preprint arXiv:2511.11181},
  year = {2025},
  url = {https://arxiv.org/abs/2511.11181},
}

@article{wang2025reliablepriors,
  title = {{Multimodal Video Emotion Recognition with Reliable Reasoning Priors}},
  author = {Zhepeng Wang and Yingjian Zhu and Guanghao Dong and Hongzhu Yi and Feng Chen and Xinming Wang and Jun Xie},
  journal = {arXiv preprint arXiv:2508.03722},
  year = {2025},
  url = {https://arxiv.org/abs/2508.03722},
}

@article{xie2025teamofone,
  title = {{Team of One: Cracking Complex Video QA with Model Synergy}},
  author = {Jun Xie and Zhaoran Zhao and Xiongjun Guan and Yingjian Zhu and Hongzhu Yi and Xinming Wang and Feng Chen and Zhepeng Wang},
  journal = {arXiv preprint arXiv:2507.13820},
  year = {2025},
  url = {https://arxiv.org/abs/2507.13820},
}

@article{xing2025lcgc,
  title = {{LCGC: Learning from Consistency Gradient Conflicting for Class-Imbalanced Semi-Supervised Debiasing}},
  author = {Weiwei Xing and Yue Cheng and Hongzhu Yi and Xiaohui Gao and Xiang Wei and Xiaoyu Guo and Yuming Zhang and Xinyu Pang},
  journal = {arXiv preprint arXiv:2504.06544},
  year = {2025},
  url = {https://arxiv.org/abs/2504.06544},
}

@inproceedings{wang2026mralign,
  title = {{MR}-{ALIGN}: Meta-Reasoning Informed Factuality Alignment for Large Reasoning Models},
  author = {Wang, Xinming and Xu, Jian and Yu, Bin and Lian, Sheng and Chen, Yi and Wang, Boran and Zhu, Yingjian and Yi, Hongzhu and Yang, Hong-Ming and Hu, Han and Liu, Cheng-Lin and Zhang, Xu-Yao},
  booktitle = {Findings of the {A}ssociation for {C}omputational {L}inguistics: {ACL} 2026},
  year = {2026},
  publisher = {Association for Computational Linguistics},
  pages = {4191--4214},
  doi = {10.18653/v1/2026.findings-acl.204},
  url = {https://aclanthology.org/2026.findings-acl.204/},
}

@inproceedings{yu2026evidbench,
  title = {When Seeing Is Not Believing{---}A Benchmark for Search-Grounded Video Misinformation Detection},
  author = {Yu, Tao and Yang, Yujia and Chai, Shenghua and Zhang, Jinshuai and Jin, Haopeng and Wang, Hao and Zhang, Minghui and Luo, Zhongtian and Long, Yuchen and Chen, Xinlong and Yang, Jiabing and Kang, Zhaolu and Zhou, Yuxuan and Cao, Siyi and Wu, Yuqing and Wang, Peifan and Dang, Tianyang and Guo, Minghui and Wang, Hanyu and Huang, Hai and Man, Zhengyu and Yang, Xingyu and Sun, Xinya and Wang, Xinming and Yi, Hongzhu and He, Zheqi and Yang, Xi and Huang, Yan and Wang, Liang},
  booktitle = {Findings of the {A}ssociation for {C}omputational {L}inguistics: {EMNLP} 2026},
  year = {2026},
  publisher = {Association for Computational Linguistics},
  pages = {16076--16118},
  doi = {10.18653/2026.findings-emnlp.763},
  url = {https://aclanthology.org/2026.findings-emnlp.763/},
}

@inproceedings{yu2026shotfinder,
  title = {{S}hot{F}inder: Imagination-Driven Open-Domain Video Shot Retrieval via Web Search},
  author = {Yu, Tao and Jin, Haopeng and Wang, Hao and Chai, Shenghua and Yang, Yujia and Luo, Zhongtian and Gong, Junhao and Guo, Jiaming and Zhang, Minghui and Chen, Xinlong and Zhang, Zhenghao and Zhou, Yuxuan and Gong, Yanpei and Liu, YuanCheng and Ding, Yiming and Zeng, Kangwei and Yang, Pengfei and Xiong, Yufei and Zhang, Shanbin and Chengshaoxiong and Ruilin, Huang and Shuo, Li and Niu, Yuxi and Zhang, Xinyuan and Xu, Yueya and Mao, Jie and Ji, Ruixuan and Zhao, Yaru and Zhang, Mingchen and Yang, Jiabing and Liu, Jiaqi and Zhang, YiFan and Yi, Hongzhu and Wang, Xinming and Zhong, Cheng and Ma, Xiao and Zhang, Zhang and Huang, Yan and Wang, Liang},
  booktitle = {Findings of the {A}ssociation for {C}omputational {L}inguistics: {EMNLP} 2026},
  year = {2026},
  publisher = {Association for Computational Linguistics},
  pages = {15915--15944},
  doi = {10.18653/2026.findings-emnlp.756},
  url = {https://aclanthology.org/2026.findings-emnlp.756/},
}

@inproceedings{zong2026l2dir,
  title = {{L}2{D}ir: Integrating $L_2$-Norm and Directional Alignment for Unsupervised Contrastive Representation Learning in Multimodal Retrieval},
  author = {Zong, Tianyu and Dai, Rui and Yi, Hongzhu and Wang, Yuanxiang and Zhang, Zhenghao and Guan, Zhenyu and Yang, Yujia and Shi, Bingkang and Ding, Yueyang and Chu, Xiangxiang and Liu, Kaikui and Xu, Jungang},
  booktitle = {Proceedings of the 64th Annual Meeting of the {A}ssociation for {C}omputational {L}inguistics (Volume 1: Long Papers)},
  year = {2026},
  publisher = {Association for Computational Linguistics},
  pages = {34717--34740},
  doi = {10.18653/v1/2026.acl-long.1604},
  url = {https://aclanthology.org/2026.acl-long.1604/},
}

@inproceedings{shi2026fairgamer,
  title = {{FAIRGAMER}: Evaluating Social Biases in {LLM}-Based Video Game {NPC}s},
  author = {Shi, Bingkang and Huang, Jen-tse and Long, Luo and Zong, Tianyu and Yi, Hongzhu and Wang, Yuanxiang and Hu, Songlin and Zhang, Xiaodan and Yao, Zhongjiang},
  booktitle = {Proceedings of the 64th Annual Meeting of the {A}ssociation for {C}omputational {L}inguistics (Volume 1: Long Papers)},
  year = {2026},
  publisher = {Association for Computational Linguistics},
  pages = {43530--43552},
  doi = {10.18653/v1/2026.acl-long.2015},
  url = {https://aclanthology.org/2026.acl-long.2015/},
}

@article{wang2026hitchhiker,
  title = {{The Hitchhiker's Guide to Autonomous Research: A Survey of Scientific Agents}},
  author = {Xinming Wang and Jian Xu and Sheng Lian and Yi Chen and Haiyang Guo and Fei Zhu and Yuanqi Shao and Minsi Ren and Hongzhu Yi and Aslan H. Feng and Hongming Yang and Tailin Wu and Han Hu and Shiming Xiang and Xu-Yao Zhang and Cheng-Lin Liu},
  journal = {IEEE Transactions on Pattern Analysis and Machine Intelligence},
  year = {2026},
  volume = {48},
  number = {11},
  pages = {15267--15286},
  doi = {10.1109/TPAMI.2026.3721500},
}

@article{yi2025abm,
  title = {{ABM}: Adaptive bias mitigation for class-imbalanced semi-supervised learning},
  author = {Hongzhu Yi and Yue Cheng and Weiwei Xing and Xiang Wei},
  journal = {Neurocomputing},
  year = {2025},
  volume = {658},
  pages = {131617},
  doi = {10.1016/j.neucom.2025.131617},
}

@inproceedings{zong2025tncse,
  title = {{TNCSE}: Tensor Norm Constraints for Unsupervised Contrastive Learning of Sentence Embeddings},
  author = {Tianyu Zong and Bingkang Shi and Hongzhu Yi and Jungang Xu},
  booktitle = {Proceedings of the AAAI Conference on Artificial Intelligence},
  year = {2025},
  volume = {39},
  pages = {26192--26201},
  doi = {10.1609/aaai.v39i24.34816},
}

@inproceedings{xie2025zeroes,
  title = {{ZeroES}: Zero-Shot Ensemble for Open-Vocabulary Video Emotion Recognition with Large Multimodal Models},
  author = {Jun Xie and Xiaohui Fan and Zhenghao Zhang and Feng Chen and Hongzhu Yi and Yingjian Zhu and Xiongjun Guan and Xinming Wang and Yue Bi and Tao Zhang and Zhepeng Wang},
  booktitle = {Proceedings of the 3rd International Workshop on Multimodal and Responsible Affective Computing},
  publisher = {ACM},
  year = {2025},
  doi = {10.1145/3746270.3760218},
}

@misc{zong2026tncme,
  title = {{TNCME}: Tensor's Norm Constraints for Unsupervised Contrastive Learning of Multimodal Embeddings},
  author = {Tianyu Zong and Hongzhu Yi and Yuanxiang Wang and Zhenghao Zhang and Yujia Yang and Zhenyu Guan and Jungang Xu},
  year = {2026},
  howpublished = {OpenReview manuscript},
  note = {Withdrawn submission to ICLR 2026},
  url = {https://openreview.net/forum?id=TkgB6sFoE2},
}

@misc{yi2026poer,
  title = {Policy Optimization with Experience Replay: Guiding Reasoning Models to Complete the Reasoning Path},
  author = {Hongzhu Yi and Xinming Wang and Zhenghao Zhang and Tianyu Zong and Yuanxiang Wang and Jun Xie and Tao Yu and Zhepeng Wang and Kaixin Xu and Feng Chen and Jiahuan Chen and Yujia Yang and Zhenyu Guan and Jungang Xu},
  year = {2026},
  howpublished = {OpenReview manuscript},
  note = {Withdrawn submission to ICLR 2026},
  url = {https://openreview.net/forum?id=AoCqLYhTD8},
}
